\documentclass[twoside]{article}

\usepackage[accepted]{aistats2023}
\usepackage{microtype}
\usepackage{graphicx}
\usepackage{subfigure}
\usepackage{booktabs}
\usepackage{setspace}
\usepackage{multirow}
\usepackage{color}

\usepackage{hyperref}
\usepackage[ruled,lined,noend,linesnumbered]{algorithm2e}
\usepackage[round]{natbib}

\input{./Definitions.tex}

\begin{document}

\renewcommand{\thefootnote}{\fnsymbol{footnote}}
%

%
\runningauthor{Qingzhong Ai, Pengyun Wang, Lirong He, Liangjian Wen, Lujia Pan, Zenglin Xu}

\twocolumn[
\aistatstitle{Generative Oversampling for Imbalanced Data via Majority-Guided VAE}
\aistatsauthor{Qingzhong Ai$^{1}$ \And Pengyun Wang$^{2,}\footnotemark[1]$ \And  Lirong He$^{1}$ \And Liangjian Wen$^{2}$ }
\onehalfspacing
\aistatsauthor{Lujia Pan$^{2}$ \And Zenglin Xu$^{3,4,}\footnotemark[2]$}

\aistatsaddress{$^{1}$University of Electronic Science and Technology of China\quad $^{2}$Noah’s Ark Lab, Huawei        Technologies \\ 
                $^{3}$Harbin Institute of Technology, Shenzhen\quad $^{4}$Peng Cheng Lab\\
                \texttt{\{qzai,lirong\_he\}@std.uestc.edu.cn}\quad \texttt{wangpyun1203@gmail.com} \\ 
                \texttt{\{wenliangjian1,panlujia\}@huawei.com}\quad \texttt{xuzenglin@hit.edu.cn}
                }

]
\footnotetext[1]{Work done while at Huawei.}
\footnotetext[2]{Corresponding Author.}
\renewcommand{\thefootnote}{\arabic{footnote}}

\begin{abstract}
    Learning with imbalanced data is a challenging problem in deep learning. Over-sampling is a widely used technique to re-balance the sampling distribution of training data. However, most existing over-sampling methods only use intra-class information of minority classes to augment the data but ignore the inter-class relationships with the majority ones, which is prone to overfitting, especially when the imbalance ratio is large. To address this issue, we propose a novel over-sampling model, called Majority-Guided VAE~(MGVAE), which generates new minority samples under the guidance of a majority-based prior. In this way, the newly generated minority samples can inherit the diversity and richness of the majority ones, thus mitigating overfitting in downstream tasks. Furthermore, to prevent model collapse under limited data, we first pre-train MGVAE on sufficient majority samples and then fine-tune based on minority samples with Elastic Weight Consolidation~(EWC) regularization. Experimental results on benchmark image datasets and real-world tabular data show that MGVAE achieves competitive improvements over other over-sampling methods in downstream classification tasks, demonstrating the effectiveness of our method.
\end{abstract}

\section{Introduction}
Modern advanced models, such as deep neural networks~(DNNs), are driven by large-scale training data of high quality, which is usually well-designed and class-balanced. However, the distributions of real-world data tend to be more complex. For example, some classes of a dataset are difficult to access due to scarcity or privacy, resulting in a significant difference in the number of training instances from the other classes. This is typically known as the class-imbalanced~\citep{johnson2019survey, huang2019deep, Wang2022imbalance} or ``long-tailed"~\citep{mahajan2018exploring, van2018inaturalist, zhang2021deep} problem. Generally, we refer to the class with sufficient samples as the majority class, and the class with few data as the minority class. On these class-imbalanced datasets, the standard training of DNNs has been found to perform poorly~\citep{wang2017learning, dong2018imbalanced, ren2018learning, he2009learning}, especially in classification tasks. The resultant classification surface of the classifier tends to be highly skewed towards the majority class due to its dominance~\citep{das2018handling}. As a result, the prediction accuracy in the minority classes is drastically affected, and the overall generalization performance suffers.
\begin{figure*}
    \centering
    \includegraphics[width=0.95\textwidth]{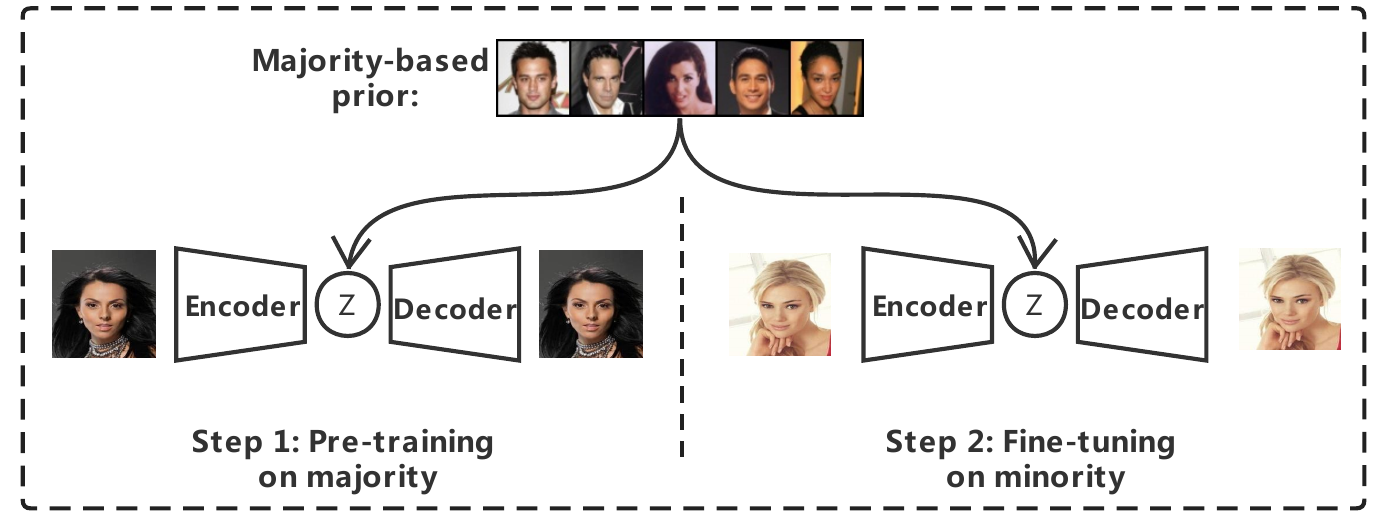}
    \caption{Pipeline of our method. Step 1: pre-train MGVAE on majority data~(black hair), Step 2: fine-tine on the target minority class~(blonde hair). Both steps under the majority-based prior.}
    \label{fig:pipeline}
\end{figure*}

To alleviate the detrimental effects of the class-imbalanced problem, one typically needs to re-balance the training objective with respect to class-wise instance size via two basic approaches: re-weighting and re-sampling. The principal concept of the re-weighting~\citep{khan2017cost, chung2015cost, lin2017focal} is to adjust the objective learning function so that the samples in the minority classes receive more attention than the majority ones. Hence, it is also called the objective-level method. Many advanced re-weighting techniques have been proposed, such as RW~\citep{huang2016learning}, CBRW~\citep{cui2019class}, FOCAL~\citep{lin2017focal}, and LDAM~\cite{cao2019learning} etc. Unlike re-weighting, re-sampling focuses on obtaining a balanced sampling distribution during training, which could be achieved by either ``under-sampling'' the majority classes~\citep{he2009learning, liu2008exploratory} or ``over-sampling'' the minority classes~\citep{cui2018large, japkowicz2000class, kang2019decoupling, wang2020devil}. So, it belongs to the data-level method.

In this paper, we focus on the technique of ``over-sampling.'' The simplest way is Random Over Sampling~(ROS)~\citep{japkowicz2000class}. As a representative algorithm, SMOTE~\citep{chawla2002smote} and its variants~\citep{han2005borderline, he2008adasyn, mullick2019generative} augment the minority class through intra-class linear interpolation sampling to alleviate overfitting, which has been widely used. However, when the number of samples in the minority class reduces to a few, the performance of SMOTE drops noticeably. Despite the recent influx of advanced over-sampling methods~\citep{fajardo2021oversampling, zhang2021learning, wang2020deep}, we notice that most of them only use intra-class information of minority classes to augment the data while ignoring the inter-class relationships, which invites effort to make further improvements. Intuitively, the majority and minority classes of the same dataset can be related, and therefore the latter can borrow the diversity and richness of the former for the purpose of a more informative data augmentation.

To this end, we propose Majority-Guided VAE~(MGVAE), a novel over-sampling model for generating the minority samples under the guidance of a majority-based prior. Specifically, we first model the distribution of the minority as a parametric distribution conditioned on the majority samples. Then, by introducing a parametric transition distribution with a hidden variable, the learning objective is tractable through variational inference, resulting in a new VAE framework. In essence, MGVAE is a generative model for minority classes with a majority-based prior. 
Nevertheless, minority samples cannot satisfy the requirement of large-scale data for the training of deep generative models to guarantee generation quality~\citep{kingma2013auto,goodfellow2014generative,dinh2014nice,ho2020denoising,ai2021bype,FanZCLXLPG22vae}. Limited data can lead to overfitting or even model collapse. To fix this flaw, we take inspiration from the paradigm of Few Shot Generation~(FSG)~\citep{li2020few, ojha2021few, wang2018transferring} and adopt a pre-training and fine-tuning framework, as shown in Figure~\ref{fig:pipeline}. In detail,  we first pre-train MGVAE on the majority's samples, which are sufficient for training a VAE from sketch. Then, we move on to fine-tune the pre-trained model based on the minority data. In addition, to prevent catastrophic forgetting of the majority and overfitting to the minority, we adopt the Elastic Weight Consolidation~(EWC) to regulate the fine-tuning progress, which proves to be beneficial. The resultant model can be used to generate new minority data by: 1) first drawing points at random from the majority based prior in the latent space; 2) and then transforming them through the learned decoder to obtain new minority data.

As we will see, the over-sampling of minority classes is straightforward with a trained MGVAE. It is able to perform one-to-one ``translation" of the instances from majority to minority, resulting in a class-balanced training dataset. To evaluate the over-sampling quality of MGVAE, we conduct extensive experiments based on various classifier backbones and imbalanced datasets. Compared to previous over-sampling techniques, our model performs the best in all cases.

\section{Methods}
\label{sec:methods}
In the big picture, we consider a classification problem on a class-imbalanced dataset $\mathcal{D}_{imb}$. Note that our method is orthogonal to the downstream tasks and thus can be seamlessly integrated into any downstream tasks without multiple training. 
For a clear elaboration, we will introduce our method by confining attention to a binary dataset with a majority class $\mathcal{X}^+ \equiv \{\x_n^+\}_{n=1}^{N^+}$ and a minority class $\mathcal{X}^- \equiv \{\x_n^-\}_{n=1}^{N^-}$. Generalizing to multi-class is straightforward and will be discussed later. In the setting of class-imbalanced, we have $N^+ \gg N^-$, where $N^+$ and $N^-$ denote the sample size of the majority and minority. Throughout this paper, vectors are symbolized by bold lowercase letters whose subscripts indicate their order, and matrices are denoted by upper-case letters.

\subsection{Majority-Guided VAE}
In general, sample size correlates with diversity. Compared to the minority class, the majority has a large-scale sample size, which means better diversity and richer information. We argue that it is crucial to be able to borrow the knowledge of the majority when augmenting the minority. To this end, we formulate the distribution of the minority as a parametric distribution conditioned on the majority. Formally, the log density distribution can be expressed as
\begin{align}
    \log p(\x^- \mid \mathcal{X}^+, \Psi),
\end{align}
where $\x^-$ denotes the minority sample,  $\mathcal{X}^+ \equiv \{\x_n^+\}_{n=1}^{N^+}$ is the set of majority samples, and $\Psi$ is the distribution parameter. More specifically, we define a parametric transition distribution $T_{\Psi}(\x^- \mid \x^+)$, which stochastically transforms a majority sample $\x^+$ into a new minority observation $\x^-$. Then, we have
\begin{align}
    \log p(\x^- \mid \mathcal{X}^+, \Psi) = \log \sum_{n=1}^{N^+} \frac{1}{N^+} T_{\Psi}(\x^- \mid \mathbf{x}^+_n),
\end{align}
where the prior probability is uniform over the majority sample. Choosing a suitable transition distribution is crucial for better approximating the data distribution, especially for a small number of observations. For example, the Kernel Density Estimator~(KDE) can be seen as a simple non-parametric transition distribution, but with limited expressive power. For the pursuit of powerful expressive ability, we introduce a parameter transition distribution with the latent variable $\z$, which defined as
\begin{align}
    T_{\Psi}(\x^- \mid \x^+) = \int_{z} r_{\phi}(\z \mid \mathbf{x}^+)p_{\theta}(\x^- \mid \z) \mathbf{d}\z,
    \label{trans_distribution}
\end{align}
where the parameters $\Psi=\{\theta,\phi\}$ and $r_{\phi}(\z \mid \mathbf{x}^+)$ is a prior based on majority. We assume that a minority observation $\x^-$ is independent to a majority sample $\x^+$ conditional on $\z$ for simplifying the formulation and optimization. In fact, a generative model of the minority is already embedded in the transition distribution in Eq.~(\ref{trans_distribution}). $r_{\phi}(\z \mid \mathbf{x}^+)$ is a majority-based prior with parameter $\phi$ for generating latent code $\z$ from a majority sample $\x^+$. Then, a decoder $p_{\theta}(\x^- \mid \z)$ with parameter $\theta$ generates new minority sample $\x^-$ from $\z$. Next, we can optimize the evidence lower bound~(ELBO) derived by variational inference as follows,
\begin{align}
    \notag
    \log&\ p(\x^- \mid \mathcal{X}^+,\Psi) \\
    \notag
    &= \log \sum_{n=1}^{N^+} \frac{1}{N^+} T_{\Psi}(\x^- \mid \x^+_n) \\
    \notag
    &= \log \sum_{n=1}^{N^+} \frac{1}{N^+} \int_{z} r_{\phi}(\z \mid \x_n^+)p_{\theta}(\x^- \mid \z) \mathbf{d}\z \\
    \notag
    &\overset{\text{VAE}}\ge \mathbb{E}_{q_{\phi}(\z \mid \x^-)} \log p_{\theta}(\x^- \mid \z) \\
    \notag
    & - \mathbb{E}_{q_{\phi}(\z \mid \x^-)} \log \frac{q_{\phi}(\z \mid \x^-)}{\sum_{n=1}^{N^+} r_{\phi}(\z \mid \x_n^+) / {N^+}} \\
    &\equiv O(\Psi, \mathcal{X}^{+}; \x^-),
    \label{objective}
\end{align}
where $O(\Psi,\mathcal{X}^{+};\x^-)$ is the optimization objective of our proposed model, MGVAE. See detailed derivation in the supplementary materials~\ref{der_obj}. According to Eq.~(\ref{objective}), we know that MGVAE is a generative model with a VAE-like framework for the minority, where the approximate posterior $q_{\phi}(\z \mid \x^-)$ is the encoder and $p_{\theta}(\x^- \mid \z)$ is the decoder. Similar to the standard VAE, the first part of the objective is the reconstruction error. The difference lies in the KL divergence of the second part, where the prior of MGVAE is a mixture prior $p(\z \mid \mathcal{X}^+)=\sum_{n=1}^{N^+} r_{\phi}(\z \mid \x_n^+) / {N^+}$, in which each component is conditioned on a majority sample.
To generate a new observation from the pre-trained MGVAE, we need both the decoder network $p_{\theta}$ and the prior network $r_{\phi}$. The generating process of MGVAE is summarised in Algorithm~\ref{MGVAE_G}. As shown, MGVAE can achieve a one-to-one probabilistic mapping from majority to minority, resulting in a class-balanced dataset.

\begin{algorithm}[!htbp]
    \caption{The Generating Process of MGVAE.}
    \label{MGVAE_G}
    \LinesNotNumbered
    \SetKwInOut{Input}{Input}
    \SetKwInOut{Output}{Output}
    \Input{ The majority samples $\mathcal{X}^+$, the decoder $p_{\theta}$, and the prior $r_{\phi}$.}
    \Output{ A generated observation $\x^-$.}
    
    $\textbf{Step.1}$  Sample $n \sim \text{Unifrom}(0, N-1)$ for obtaining a random sample $\x_n^+$ from the majority $\mathcal{X}^+$.
    
    $\textbf{Step.2}$  Sample $\z \sim r_{\phi}(\cdot \mid \x^+_n)$ using the majority point based prior $r_{\phi}$ to obtain a latent code $\z$.
    
    $\textbf{Step.3}$  Sample $\x^- \sim p_{\theta}(\cdot \mid \z)$ using the decoder $p_{\theta}$ for generating a new observation $\x^-$.
\end{algorithm}

\paragraph{Implemental Details} In practice, following advanced works, such as VampPrior, Exemplar VAE, and ByPE-VAE, the prior $r_{\phi}$ is modeled as a Gaussian distribution $\mathcal{N}(\z \mid \mu_{\phi}(\x), \sigma^2 I)$, in which the parametric mean function $\mu_{\phi}$ is shared with encoder $q$, and the covariance function is an isotropic Gaussian with a scalar parameter $\sigma$. Therefore, the parameters in $r_{\phi}$ can be updated every iteration by the backpropagation gradient descent. Moreover, instead of using entire majority samples, we randomly down-sample a fix-sized majority in each step to compute the prior for computation efficiency. The detailed training process is summarized in Algorithm~\ref{alg:training_process} in supplementary materials~\ref{appsec:training process}.

\paragraph{Generalization to multi-class}It is straightforward to generalize MGVAE to the multiple class-imbalanced classification problems, such as long-tailed learning. For a long-tailed dataset $\mathcal{D}_{LT}$ with $K$ classes $\{C_{i}\}_{i=1}^{K}$, we have $N_1>N_2>... >N_K$, where $N_i$ is the sample size of class $C_i$. To over-sampling the dataset $\mathcal{D}_{LT}$ with MGVAE, we construct $K-1$ binary class datasets. The majority class of each binary dataset is $C_1$, while the minority is $C_2-C_K$, respectively. After that, we can obtain $K-1$ MGVAEs, each corresponding to one minority class. Finally, the over-sampling of the dataset $\mathcal{D}_{LT}$ is completed by randomly sampling each MGVAE according to Algorithm~\ref{MGVAE_G}, resulting in a balanced dataset.

\subsection{Training with limited data}
\begin{figure}[htbp]
    \centering

    \subfigure[$N^-=3000$]{
        \includegraphics[width=0.26\linewidth]{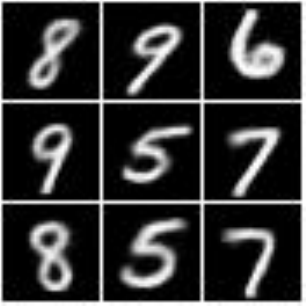}
        \label{fig:example_mnist_a}
    }
    \hspace{-8pt}
    \subfigure[$N^-=300$]{
        \includegraphics[width=0.26\linewidth]{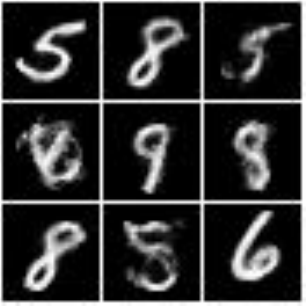}
    }
    \hspace{-8pt}
    \subfigure[$N^-=50$]{
        \includegraphics[width=0.26\linewidth]{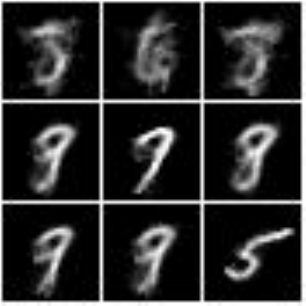}
        \label{fig:example_mnist_c}
    }
    \subfigure[$N^-=300$]{
        \includegraphics[width=0.26\linewidth]{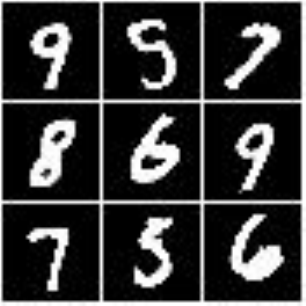}
        \label{fig:example_mnist_d}
    }
    \hspace{-8pt}
    \subfigure[$N^-=50$]{
        \includegraphics[width=0.26\linewidth]{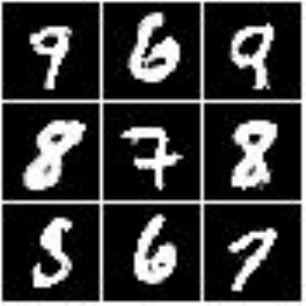}
    }
    \hspace{-8pt}
    \subfigure[$N^-=10$]{
        \includegraphics[width=0.26\linewidth]{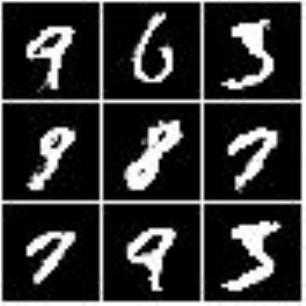}
        \label{fig:example_mnist_f}
    }
    \caption{The sampling results of MGVAE. (a-c): Without the Pre-training and Fine-tuning; (d-f): With the Pre-training and Fine-tuning. The label under each figure is the size of the downsampled minority samples.}
    \label{fig:example_mnist}

\end{figure}

So far, we have obtained a generative model and corresponding generating process for the minority with a majority-based prior. Unfortunately, the model does not work well yet to generate samples with limited data. We test MGVAE on modified-MNIST, where all 0-4 are used as the majority and downsampled 5-9 as the minority. The sampling results are shown in Figure~\ref{fig:example_mnist_a} -~\ref{fig:example_mnist_c}, from which we can find that as the size of minority samples decreases, generation quality gradually deteriorates until indistinguishable. We argue this is a generative dilemma for all modern deep generative models. When the number of observations is much smaller than model parameters, the model tends to overfit or collapse.

\subsubsection{Pre-training and Fine-tuning}
To address this issue, we adopt the process of pre-training and fine-tuning, which is borrowed from the idea of FSG. Specifically, we first pre-train an MGVAE based on the majority samples, which is adequate for the training from sketch. After convergence, we move on to the minority, fine-tuning the pre-trained model based on limited minority samples. The pipeline of the training process is shown in Figure~\ref{fig:pipeline}. Likewise, we can sample the model after the training process of ``pre-training and fine-tuning" according to Algorithm~\ref{MGVAE_G}. The sampling results are shown in Figure~\ref{fig:example_mnist_d} -~\ref{fig:example_mnist_f}. By comparing with Figure~\ref{fig:example_mnist_a} -~\ref{fig:example_mnist_c}, we find that the samples generated by the current model are meaningful and distinguishable, even when the size of the minority sample is extremely small~(e.g., 10). More visual results are presented in the experimental section. 

\subsubsection{EWC regularization}
Further, we employ Elastic Weight Consolidation~(EWC) regularization during the fine-tuning process. EWC is a commonly used fine-tune technique, first proposed in \cite{kirkpatrick2017overcoming} to avoid catastrophic forgetting in continuous learning, and \cite{li2020few} introduces EWC to the FSG setting. In \cite{nguyen2017variational}, EWC is introduced for VAE in Variational Continual Learning. To prevent overfitting to the new domain and the catastrophic forgetting of the old one, EWC preserves the important parameters by penalizing parameter changes. The importance of each parameter is measured by Fisher Information, defined as follows,
\begin{align}
    \label{eq:fisher}
    F=\mathbb{E}\left[-\frac{\partial^{2}}{\partial \Psi^{2}} \mathcal{L}\left(X \mid \Psi\right)\right],
\end{align}
where $\mathcal{L}$ is the log-likelihood function that can be approximated by ELBO in VAE, $X$ is a collection of generated data, and $\Psi$ is the learned parameter value. Given the model pre-trained in the majority with learned parameters $\Psi_{+}$, we could get the Fisher Information vector $F$ and then formalize the fine-tuning process as
\begin{align}
    \label{eq:final}
    O_{\text{EWC}} = O(\Psi,\mathcal{X}^{+};\x^-) + \lambda \sum_{i} F_{i}\left(\Psi_{i}-\Psi_{+, i}\right)^{2},
\end{align}
where $O(\Psi,\mathcal{X}^{+};\x^-)$ is the optimization objective in Eq.~(\ref{objective}), $\lambda$ is the regularization weight, and $i$ is the index of each model parameter. Note that $O_{\text{EWC}}$ is our final optimization objective function.

\section{Experiments}
To evaluate the over-sampling quality of MGVAE, we consider classification as our downstream evaluation task. 
Given an imbalanced dataset $\mathcal{D}_{imb}$, we first augment the dataset by MGVAE to obtain a balanced dataset $\mathcal{D}_{bal}$. Then, we train the classifier based on the augmented dataset $\mathcal{D}_{bal}$. We test our model on several datasets of different scales and data types, including image datasets MNIST~\citep{lecun1998gradient}, FashionMNIST~\citep{xiao2017fashion}, CelebA~\citep{liu2015faceattributes}, and several tabular datasets, namely Musk~\citep{asuncion2007uci}, Water Quality, and Isolet. We use different network architectures of MGVAE depending on the dimensionality and size of the dataset. The same goes for the classifier. For the image data, we also give the corresponding visual results, aiming to better demonstrate the significance and validity of our method. Finally, some analysis experiments, including ablation study and sensitivity analysis, are conducted to understand the proposed method in detail.

\paragraph{Baseline Methods.}
We adopt a wide range of existing methods as our baselines, as follows:
\begin{enumerate}
    \item \textbf{Empirical risk minimization~(ERM)}: the standard training method with cross-entropy loss and without any balancing operation.
    \item \textbf{Random over-sampling~(ROS)}~\citep{japkowicz2000class}: balancing the sampling distribution by repeated random sampling of the minority class. 
    \item \textbf{SMOTE}~\citep{chawla2002smote}: balancing the sampling distribution by linear interpolating nearest neighbors in the minority class.
    \item \textbf{Re-weighting~(RW)}~\citep{huang2016learning}: modifying the objective function according to the class sample size.
    \item \textbf{Class-balanced re-weighting~(CBRW)}~\citep{cui2019class}: an improved variant of RW, introducing the effective sample number $E_k = (1-\beta^{N_k}) / (1-\beta)$ for each class, where $\beta$ is set to 0.9999.
    \item \textbf{FOCAL}~\citep{lin2017focal}: aiming to balance the sample-wise classification loss for model training by down-weighing the well-classiﬁed samples.
    \item \textbf{LDAM}~\citep{cao2019learning}: a label-distribution-aware margin loss that encourages few-shot classes to have larger margins.
    \item \textbf{OCVAE}~\citep{fajardo2021oversampling}: \textbf{O}ver-sampling the minority by a \textbf{C}onditional \textbf{VAE}, a generative model augmenting the dataset conditioned on the class label.
    \item \textbf{OCGAN} or \textbf{OCDCGAN}~\citep{fajardo2021oversampling}: \textbf{O}ver-sampling the minority by \textbf{C}onditional \textbf{GAN}, another generative model augmenting the dataset conditioned on the class label.
\end{enumerate}
Note that the main network architecture of OCVAE, OCGAN, and OCDCGAN is consistent with our model for fairness. To distinguish all comparison methods more clearly, we classify them into four categories, classified as traditional over-sampling methods~(ROS, SMOTE); re-weighting methods~(RW, CBRW); other loss functions~(FOCAL, LDAM), and DGMs-based over-sampling methods~(OCVAE, OCGAN, and OCDCGAN).

\paragraph{Datasets Information.}
We test our model on both benchmark image datasets and real-world tabular data. For the image dataset, we chose MNIST, FashionMNIST, and CelebA, all of which are artificial class-balanced datasets. Therefore, we need to downsample the dataset to satisfy the class-imbalance setting. Similar to \cite{fajardo2021oversampling}, we conduct a challenging binary classification task for MNIST and FashionMNIST. Taking MNIST as an example, we take all classes of 0-4 as the majority classes with size 30000=6000*5, and downsample classes of 5-9 to 300=60*5 and 50=10*5 as the minority class. i.e., the imbalance ratio~(IR) is $\rho=100$ and $\rho=600$, respectively. The processing of FashionMNIST is similar. For CelebA, we chose hair color as the label to form a five-class long-tail dataset, where the sizes of black hair, blonde hair, blad hair, brown hair, and gray hair are 20000, 10000, 2500, 1000, and 200, respectively. Black hair is used as the majority class. In addition, we also conducted experiments on several real-world tabular datasets, which have natural class imbalance and therefore do not require additional manipulations. Information of all datasets is summarised in Table~\ref{tab:data_info}.

\begin{table}[!htbp]
\begin{center}
\caption{Data Information. ``Maj" stands for majority, ``Min" stands for minority, and ``c" stands for class. The number X in ``\textit{Dataset}-X" stands for imbalance ratio. }
\label{tab:data_info}
\resizebox{\linewidth}{!}{
    \begin{tabular}{cc|c}
    \toprule[2pt]
    \multicolumn{2}{c|}{Datasets}                                    & Split Way \\ \hline
    \multicolumn{1}{l|}{\multirow{5}{*}{Images}}  & MNIST-100        & Maj: 0-4~(30000); Min: 5-9~(300) \\ \cline{2-3} 
    \multicolumn{1}{l|}{}                         & MNIST-600        & Maj: 0-4~(30000); Min: 5-9~(50)  \\ \cline{2-3} 
    \multicolumn{1}{l|}{}                         & FashionMNIST-100 & Maj: c0-c4~(30000); Min: c5-c9~(300) \\ \cline{2-3} 
    \multicolumn{1}{l|}{}                         & FashionMNIST-600 & Maj: c0-c4~(30000); Min: c5-c9~(50)  \\ \cline{2-3} 
    \multicolumn{1}{l|}{}                         & CelebA-100       & Long tail: 20000:10000:2500:1000:200         \\ \hline
    \multicolumn{1}{l|}{\multirow{3}{*}{Tabular}} & Musk-6.6         & Maj:c0~(5381); Min: c1~(817)       \\ \cline{2-3} 
    \multicolumn{1}{l|}{}                         & Water Quality-11.1   & Maj: c0~(6784); Min: c1~(612)          \\ \cline{2-3} 
    \multicolumn{1}{l|}{}                         & Isolet-17.5      & Maj: c0~(6997); Min: c1~(400)          \\ \hline
    \end{tabular}
    }
\end{center}
\end{table}

\paragraph{Evaluation Metrics.}
We use three evaluation metrics to assess the classification performance of all methods under the same balanced test distribution, namely Balanced Accuracy~(B-ACC)~\citep{huang2016learning, wang2017learning, kim2020m2m}, Average Class Speciﬁc Accuracy~(ACSA)~\citep{huang2016learning, wang2017learning, mullick2019generative}, and Geometric Mean~(GM)~\citep{kubat1997addressing, branco2016survey}. B-ACC is same as the standard accuracy metric on the balanced dataset. The other two metrics are not biased toward any particular class, therefore more suitable for evaluating the performance under an imbalanced setting. Note that all numerical results in this section average three random trials.

\subsection{Results}
We evaluate the performance of all methods in the following two aspects: the numerical results on classification as a quantitative comparison, and the other is the visual qualitative results. All results tables highlight the best results in bold, and the second-best results are underlined.

\subsubsection{Quantitative Results}
\paragraph{MNIST and FashionMNIST.} Due to the similarity of the dataset scale~(28*28), the experimental setups of MNIST and FashionMNIST are basically the same. We adopt two imbalance ratios, $\rho=100$ and $\rho=600$, for each dataset. The classifier is a 2-layer fully-connected neural network with 256 and 128 middle nodes, trained for 100 epochs with mini-batch size 100. The learning rate is initialized to 1$e$-3 and decreases with an exponential schedule of $\gamma = 0.95$. Besides, the architecture of all generative models, including MGVAE and OCVAE, remains consistent to ensure fairness. The EWC regularization weight $\lambda$ in MGVAE is selected from the set of candidate $\{5e2,5e4,5e6,5e8\}$. Notice that we omit the results of OCGAN due to the inability to train properly on small-scale data.
\begin{table}[!htbp]
\begin{center}
\caption{Comparison of classification performance on MNIST with two imbalance radios.}
\label{tab:mnist}
\resizebox{\linewidth}{!}{
    \begin{tabular}{c|cll|cll}
    \toprule[2pt]
    IR & \multicolumn{3}{c|}{$\rho = 100$} & \multicolumn{3}{c}{$\rho=600$} \\ \hline
    Methods         & \multicolumn{1}{c}{B-ACC} & \multicolumn{1}{c}{ASCA} & \multicolumn{1}{c|}{GM} & \multicolumn{1}{c}{B-ACC} & \multicolumn{1}{c}{ASCA} & \multicolumn{1}{c}{GM} \\ \hline
    
    ERM & \multicolumn{1}{c}{$51.4\pm 0.0$} & \multicolumn{1}{c}{$50.0 \pm 0.0$} & \multicolumn{1}{c|}{$0.0 \pm 0.0$} & \multicolumn{1}{c}{$51.4 \pm 0.0$} & \multicolumn{1}{c}{$50.0 \pm 0.0$} & \multicolumn{1}{c}{$0.0 \pm 0.0$}\\
    FOCAL & \multicolumn{1}{c}{$51.4\pm 0.0$} & \multicolumn{1}{c}{$50.0 \pm 0.0$} & \multicolumn{1}{c|}{$0.0 \pm 0.0$} & \multicolumn{1}{c}{$51.4 \pm 0.0$} & \multicolumn{1}{c}{$50.0 \pm 0.0$} & \multicolumn{1}{c}{$0.0 \pm 0.0$}\\
    RW  & \multicolumn{1}{c}{$77.4 \pm 1.2$} & \multicolumn{1}{c}{$76.7 \pm 1.2$} & \multicolumn{1}{c|}{$73.1 \pm 1.3$} & \multicolumn{1}{c}{$59.9 \pm 1.6$} & \multicolumn{1}{c}{$58.8 \pm 1.4$} & \multicolumn{1}{c}{$41.7 \pm 1.8$}\\
    CBRW  & \multicolumn{1}{c}{$75.1 \pm 0.8$} & \multicolumn{1}{c}{$74.3 \pm 0.7$} & \multicolumn{1}{c|}{$69.8 \pm 1.3$} & \multicolumn{1}{c}{$56.1 \pm 0.5$} & \multicolumn{1}{c}{$55.1 \pm 0.3$} & \multicolumn{1}{c}{$31.2 \pm 1.2$}\\
    LDAM  & \multicolumn{1}{c}{$82.9 \pm 0.5$} & \multicolumn{1}{c}{$82.4 \pm 0.6$} & \multicolumn{1}{c|}{$80.3 \pm 0.7$} & \multicolumn{1}{c}{$63.1 \pm 0.9$} & \multicolumn{1}{c}{$62.0 \pm 0.8$} & \multicolumn{1}{c}{$48.7 \pm 1.0$}\\
    RS  & \multicolumn{1}{c}{$79.2 \pm 0.3$} & \multicolumn{1}{c}{$78.5 \pm 0.3$} & \multicolumn{1}{c|}{$75.4 \pm 0.2$} & \multicolumn{1}{c}{$58.5\pm 1.0$} & \multicolumn{1}{c}{$57.3 \pm 1.1$} & \multicolumn{1}{c}{$37.7 \pm 2.0$}\\
    SMOTE  & \multicolumn{1}{c}{$80.6 \pm 0.3$} & \multicolumn{1}{c}{$80.1 \pm 0.2$} & \multicolumn{1}{c|}{$77.4 \pm 0.1$} & \multicolumn{1}{c}{$60.0 \pm 0.9$} & \multicolumn{1}{c}{$58.3 \pm 1.0$} & \multicolumn{1}{c}{$40.2 \pm 1.1$}\\
    OCVAE  & \multicolumn{1}{c}{$\underline{83.0} \pm 0.4$} & \multicolumn{1}{c}{$\underline{82.6} \pm 0.4$} & \multicolumn{1}{c|}{$\underline{80.6} \pm 0.6$} & \multicolumn{1}{c}{$\underline{63.8} \pm 0.2$} & \multicolumn{1}{c}{$\underline{62.8} \pm 0.5$} & \multicolumn{1}{c}{$\underline{50.7} \pm 0.5$}\\
    \textbf{MGVAE}  & \multicolumn{1}{c}{$\textbf{85.0} \pm 0.2$} & \multicolumn{1}{c}{$\textbf{84.6} \pm 0.2$} & \multicolumn{1}{c|}{$\textbf{83.2} \pm 0.2$} & \multicolumn{1}{c}{$\textbf{65.4} \pm 1.0$} & \multicolumn{1}{c}{$\textbf{64.4} \pm 1.1$} & \multicolumn{1}{c}{$\textbf{53.4} \pm 1.1$}\\
    \bottomrule
    \end{tabular}
}
\end{center}
\end{table}

\begin{table}[!htbp]
\caption{Comparison of classification performance on FashionMNIST with two imbalance radios.}
\label{tab:fashionmnist}
\begin{center}
\resizebox{\linewidth}{!}{
    \begin{tabular}{c|cll|cll}
    \toprule[2pt]
    IR & \multicolumn{3}{c|}{$\rho = 100$} & \multicolumn{3}{c}{$\rho=600$} \\ \hline
    Methods         & \multicolumn{1}{c}{B-ACC} & \multicolumn{1}{c}{ASCA} & \multicolumn{1}{c|}{GM} & \multicolumn{1}{c}{B-ACC} & \multicolumn{1}{c}{ASCA} & \multicolumn{1}{c}{GM} \\ \hline
    
    ERM & \multicolumn{1}{c}{$86.4\pm 0.1$} & \multicolumn{1}{c}{$86.4 \pm 0.1$} & \multicolumn{1}{c|}{$85.3 \pm 0.1$} & \multicolumn{1}{c}{$80.7 \pm 0.3$} & \multicolumn{1}{c}{$80.8 \pm 0.4$} & \multicolumn{1}{c}{$78.3 \pm 0.4$}\\
    FOCAL & \multicolumn{1}{c}{$86.9\pm 0.4$} & \multicolumn{1}{c}{$86.9 \pm 0.4$} & \multicolumn{1}{c|}{$85.9 \pm 0.4$} & \multicolumn{1}{c}{$81.9 \pm 0.4$} & \multicolumn{1}{c}{$82.0 \pm 0.3$} & \multicolumn{1}{c}{$79.9 \pm 0.4$}\\
    RW  & \multicolumn{1}{c}{$\underline{87.8} \pm 0.5$} & \multicolumn{1}{c}{$\underline{87.8} \pm 0.5$} & \multicolumn{1}{c|}{$\underline{86.9} \pm 0.6$} & \multicolumn{1}{c}{$83.6 \pm 0.9$} & \multicolumn{1}{c}{$83.6 \pm 0.9$} & \multicolumn{1}{c}{$81.9 \pm 1.1$}\\
    CBRW  & \multicolumn{1}{c}{$86.9 \pm 0.2$} & \multicolumn{1}{c}{$87.0 \pm 0.3$} & \multicolumn{1}{c|}{$85.9 \pm 0.3$} & \multicolumn{1}{c}{$82.8 \pm 0.9$} & \multicolumn{1}{c}{$82.8 \pm 0.9$} & \multicolumn{1}{c}{$81.0 \pm 1.1$}\\
    LDAM  & \multicolumn{1}{c}{$87.7 \pm 0.2$} & \multicolumn{1}{c}{$87.6 \pm 0.3$} & \multicolumn{1}{c|}{$86.8 \pm 0.3$} & \multicolumn{1}{c}{$\underline{83.7} \pm 0.3$} & \multicolumn{1}{c}{$\underline{83.8} \pm 0.3$} & \multicolumn{1}{c}{$\underline{81.9} \pm 0.2$}\\
    RS  & \multicolumn{1}{c}{$87.6 \pm 0.1$} & \multicolumn{1}{c}{$87.5 \pm 0.2$} & \multicolumn{1}{c|}{$86.6 \pm 0.2$} & \multicolumn{1}{c}{$81.7 \pm 0.7$} & \multicolumn{1}{c}{$81.7 \pm 0.8$} & \multicolumn{1}{c}{$79.5 \pm 0.5$}\\
    SMOTE  & \multicolumn{1}{c}{$86.4 \pm 0.3$} & \multicolumn{1}{c}{$86.5 \pm 0.3$} & \multicolumn{1}{c|}{$85.3 \pm 0.4$} & \multicolumn{1}{c}{$80.7 \pm 0.8$} & \multicolumn{1}{c}{$81.1 \pm 0.5$} & \multicolumn{1}{c}{$78.7 \pm 0.6$}\\
    OCVAE  & \multicolumn{1}{c}{$86.7 \pm 0.3$} & \multicolumn{1}{c}{$86.7 \pm 0.3$} & \multicolumn{1}{c|}{$85.8 \pm 0.4$} & \multicolumn{1}{c}{$82.8 \pm 0.4$} & \multicolumn{1}{c}{$82.8 \pm 0.4$} & \multicolumn{1}{c}{$81.1 \pm 0.5$}\\
    \textbf{MGVAE}  & \multicolumn{1}{c}{$\textbf{88.3} \pm 0.1$} & \multicolumn{1}{c}{$\textbf{88.4} \pm 0.1$} & \multicolumn{1}{c|}{$\textbf{87.6} \pm 0.1$} & \multicolumn{1}{c}{$\textbf{84.8} \pm 0.4$} & \multicolumn{1}{c}{$\textbf{84.8} \pm 0.4$} & \multicolumn{1}{c}{$\textbf{83.6} \pm 0.4$}\\
    \bottomrule
    \end{tabular}
}
\end{center}
\end{table}

The main results for MNIST and FashionMNIST are presented in Table~\ref{tab:mnist} and Table~\ref{tab:fashionmnist}, respectively. Overall, the results show that our method consistently leads by a large margin compared to other tested methods. In particular, our method performs better than all other over-sampling methods, including RS, SMOTE, and OCVAE, which means that the samples augmented by MGVAE have much less noise. Besides, we find that the MGVAE achieves a larger lead under the imbalance ratio~(IR) $\rho=600$. For example, the percentage improvement of the GM term compared to SMOTE is $\Delta = 5.8$ and $\Delta = 13.2$, respectively. This means that our method remains effective in the case of extreme imbalance.

\paragraph{CelebA.}
Then, we move on to CelebA, a much larger multi-label celebrity face dataset with a resized resolution of 64*64*3. We downsample the dataset based on hair color to obtain a 5-class long-tailed dataset with an imbalance rate of $\rho=N1/N5=100$. For classifier selection, we train a ResNet-20~\citep{he2016identity} using cross-entropy loss for 90 epochs with mini-batch size 100. The learning rate is initialized to 0.01 with a weight decay of 2$e$-4 and an exponential decrease of $\gamma=0.95$. In addition, all the generative models, including MGVAE, OCVAE, and OCDCGAN, introduce convolution layers, which are more suitable for complex image generation. Meanwhile, the main architecture of the models is kept consistent. The EWC regularization weight $\lambda$ in MGVAE is set to 50. To pursue uniformity, we only compare MGVAE with the over-sampling-based methods in this section.

\begin{table}[!htbp]
    \caption{Comparison of classification performance on the long-tail CelebA.}
    \label{tab:celeba}
    \begin{center}
    \resizebox{0.9\linewidth}{!}{
    \begin{tabular}{c|ccc}
        \toprule[2pt]
        Methods & B-ACC & ASCA & GM \\
        \hline
        ERM & $62.3 \pm 0.9$ & $63.8 \pm 0.5$ & $40.2 \pm 0.3$ \\
        RS & $64.2 \pm 0.5$ & $65.6 \pm 0.4$ & $45.0 \pm 1.2$ \\
        SMOTE & $63.5 \pm 0.8$ & $65.0 \pm 0.6$ & $42.2 \pm 0.8$ \\
        OCVAE & $64.4 \pm 0.8$ & $65.5 \pm 1.0$ & $48.5 \pm 0.6$ \\
        OCDCGAN & $\underline{65.8} \pm 0.1$ & $\underline{67.2} \pm 0.0$& $\underline{52.4} \pm 1.0$\\
        \textbf{MGVAE~(ours)} & $\textbf{66.8} \pm 0.2$ & $\textbf{68.0} \pm 0.2$& $\textbf{55.6} \pm 0.2$\\
        \bottomrule
    \end{tabular}
    }
            
    \end{center}
\end{table}

The results are shown in Table~\ref{tab:celeba}. Once again, our method outperforms all other baseline methods, demonstrating that the minority samples augmented by MGVAE are more diverse and meaningful. Remarkably, as a VAE-based method, MGVAE surpasses OCDCGAN, further showing our algorithm's effectiveness.

\begin{table}[!t]
    \caption{Comparison of classification performance on the Musk.}
    \label{tab:musk}
    \begin{center}
    \resizebox{0.9\linewidth}{!}{
    \begin{tabular}{c|ccc}
        \toprule[2pt]
        Methods & B-ACC & ASCA & GM \\
        \hline
        ERM   & $50.0 \pm 0.0$ & $50.0 \pm 0.0$ & $ 0.0 \pm 0.0$\\
        FOCAL & $87.1 \pm 1.5$ & $87.2 \pm 1.6$ & $86.1 \pm 1.4$\\
        RW    & $85.6 \pm 1.1$ & $86.7 \pm 1.0$ & $84.3 \pm 1.2$\\
        CBRW  & $84.2 \pm 0.5$ & $84.3 \pm 0.5$ & $82.7 \pm 0.6$\\
        LDAM  & $89.7 \pm 1.2$ & $89.8 \pm 1.2$ & $89.0 \pm 1.3$\\
        RS    & $84.9 \pm 0.9$ & $85.0 \pm 1.0$ & $83.6 \pm 1.1$ \\
        SMOTE & $83.6 \pm 1.3$ & $83.5 \pm 1.4$ & $81.8 \pm 1.7$ \\
        OCVAE & $\underline{90.5} \pm 0.3$ & $\underline{90.1} \pm 0.2$ & $\underline{89.8} \pm 0.2$ \\
        
        \textbf{MGVAE~(ours)} & $\textbf{92.4} \pm 1.2$ & $\textbf{92.4} \pm 1.3$& $\textbf{92.1} \pm 1.4$\\
        \bottomrule
    \end{tabular}
    }
    \end{center}
\end{table}

\begin{table}[!t]
    \caption{Comparison of classification performance on the Water Quality.}
    \label{tab:water_quality}
    \begin{center}
    \resizebox{0.9\linewidth}{!}{
    \begin{tabular}{c|ccc}
        \toprule[2pt]
        Methods & B-ACC & ASCA & GM \\
        \hline
        ERM   & $50.0 \pm 0.0$ & $50.0 \pm 0.0$ & $ 0.0 \pm 0.0$\\
        FOCAL & $65.7 \pm 0.7$ & $65.7 \pm 0.7$ & $\underline{63.3} \pm 1.1$\\
        RW    & $57.7 \pm 0.1$ & $57.6 \pm 0.3$ & $48.7 \pm 0.2$\\
        CBRW  & $56.4 \pm 0.2$ & $56.4 \pm 0.3$ & $49.5 \pm 0.6$\\
        LDAM  & $\underline{65.8} \pm 0.3$ & $\underline{65.9} \pm 0.2$ & $62.0 \pm 0.3$\\
        RS    & $58.3 \pm 1.5$ & $58.2 \pm 1.4$ & $51.3 \pm 1.1$ \\
        SMOTE & $56.8 \pm 1.4$ & $56.0 \pm 1.5$ & $51.9 \pm 1.6$ \\
        OCVAE & $61.9 \pm 1.2$ & $62.1 \pm 1.9$ & $58.3 \pm 1.5$ \\
        
        \textbf{MGVAE~(ours)} & $\textbf{67.2} \pm 1.1$ & $\textbf{67.3} \pm 1.3$& $\textbf{65.9} \pm 1.2$\\
        \bottomrule
    \end{tabular}
    }
    \end{center}
\end{table}

\begin{table}[!t]
    \caption{Comparison of classification performance on the Isolet.}
    \label{tab:isolet}
    \begin{center}
    \resizebox{0.9\linewidth}{!}{
    \begin{tabular}{c|ccc}
        \toprule[2pt]
        Methods & B-ACC & ASCA & GM \\
        \hline
        ERM   & $50.0 \pm 0.0$ & $50.0 \pm 0.0$ & $ 0.0 \pm 0.0$\\
        FOCAL & $50.0 \pm 0.0$ & $50.0 \pm 0.0$ & $ 0.0 \pm 0.0$\\
        RW    & $92.9 \pm 0.4$ & $92.9 \pm 0.2$ & $92.7 \pm 0.2$\\
        CBRW  & $92.8 \pm 0.4$ & $92.8 \pm 0.5$ & $92.6 \pm 0.6$\\
        LDAM  & $93.5 \pm 1.0$ & $93.5 \pm 1.0$ & $93.3 \pm 1.1$\\
        RS    & $92.9 \pm 0.3$ & $92.8 \pm 0.4$ & $92.7 \pm 0.4$ \\
        SMOTE & $\underline{93.8} \pm 0.7$ & $\underline{93.7} \pm 0.8$ & $\underline{93.6} \pm 0.6$\\
        OCVAE & $91.6 \pm 0.4$ & $91.7 \pm 0.2$ & $91.5 \pm 0.2$ \\
        
        \textbf{MGVAE~(ours)} & $\textbf{95.1} \pm 0.6$ & $\textbf{95.0} \pm 0.7$& $\textbf{94.9} \pm 0.7$\\
        \bottomrule
    \end{tabular}
    }
    \end{center}
\end{table}

\begin{figure}[!t]
    \centering
    \subfigure[MNIST~(left: $\rho=600$; right: $\rho=100$)]{
        \begin{minipage}{0.47\linewidth}
            \centering
                \includegraphics[width=0.48\linewidth]{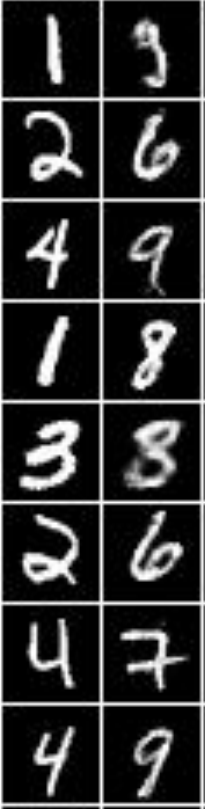}
                \includegraphics[width=0.48\linewidth]{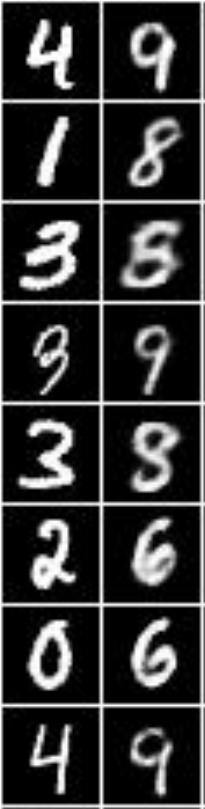}
        \end{minipage}
        \label{fig:mnist_et_vae}
    }
    \subfigure[FashionMNIST~(left: $\rho=600$; right: $\rho=100$)]{
        \begin{minipage}{0.47\linewidth}
            \centering
                \includegraphics[width=0.48\linewidth]{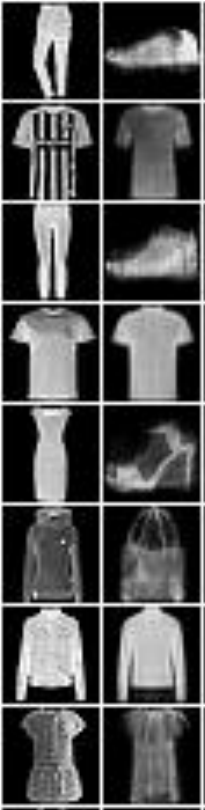}
                \includegraphics[width=0.48\linewidth]{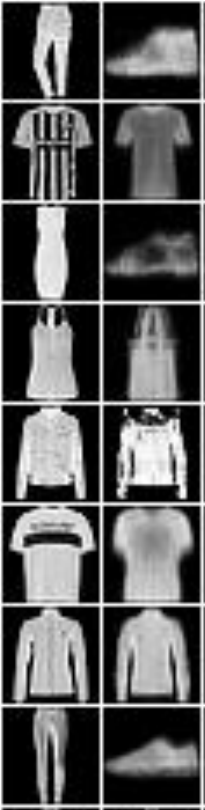}
        \end{minipage}
        \label{fig:fashionmnist_et_vae}
    }
    \subfigure[CelebA~(from left to right: black,blonde,bald,brown,and gray)]{
        \includegraphics[width=0.48\linewidth]{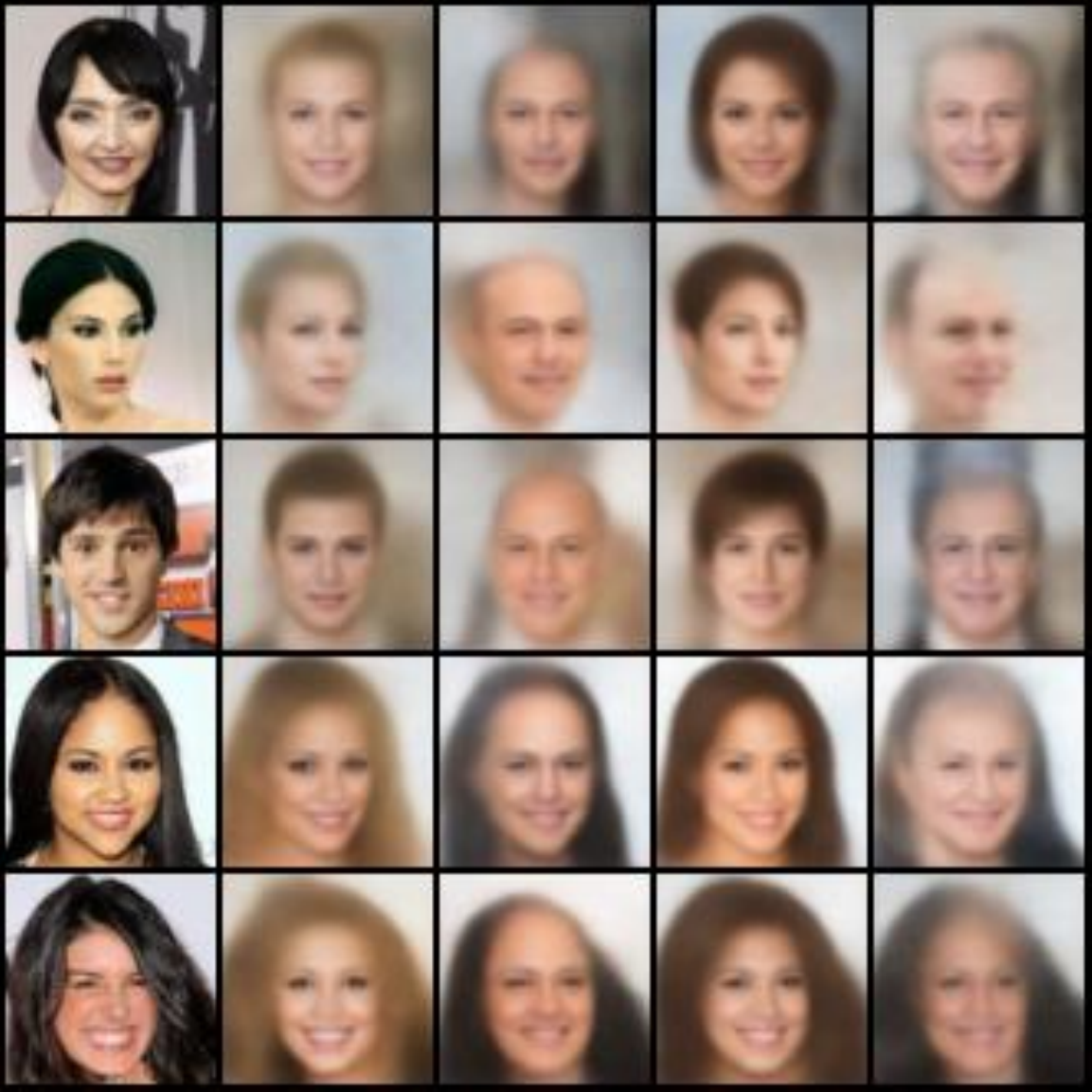}
        \includegraphics[width=0.48\linewidth]{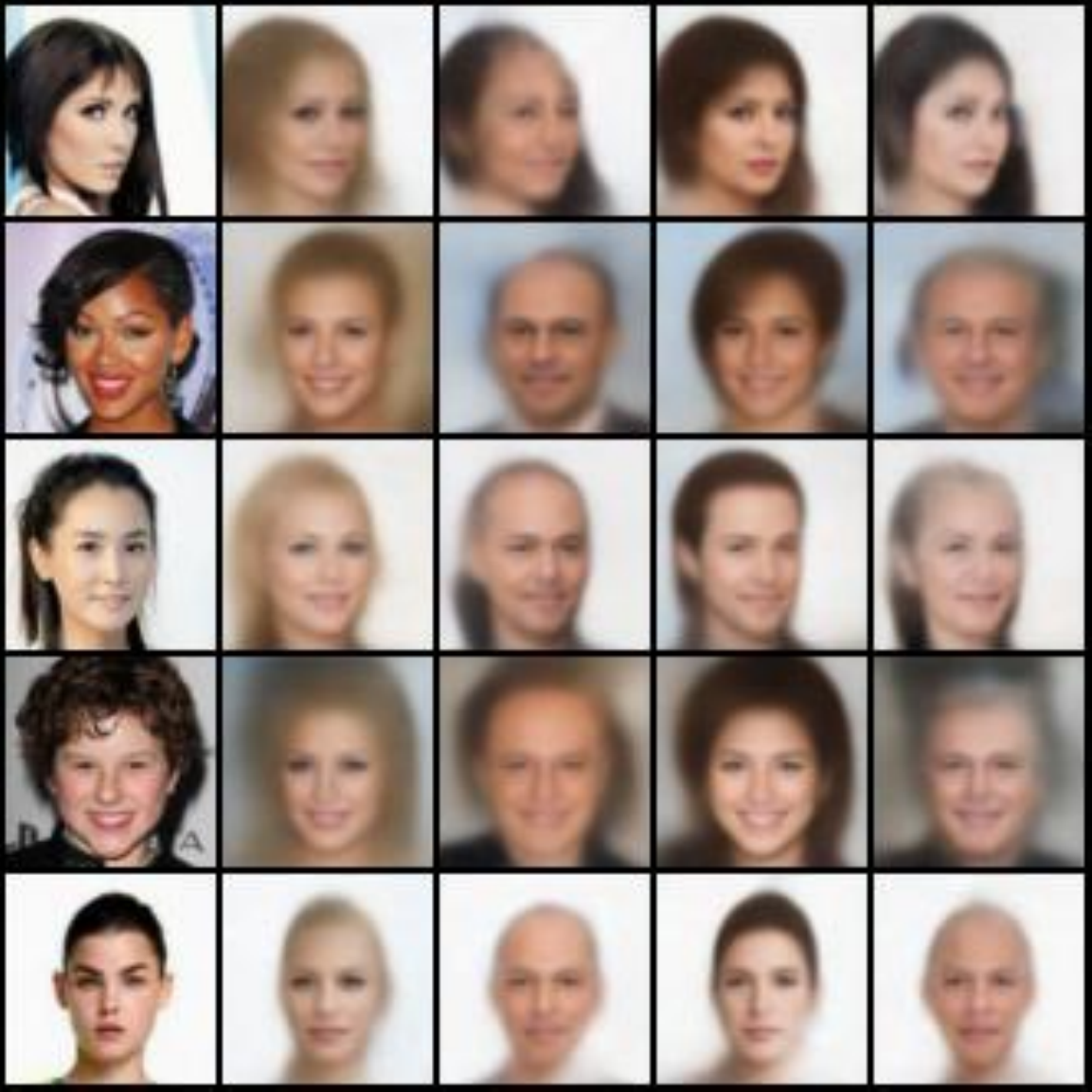}
        \label{fig:celeba_et_vae}
    }
    \caption{Samples from MGVAE. In each plate, the first column is the reference majority sample, and the rest columns are the corresponding generated minority ones. Best viewed in color.}
    \label{fig:exemplar_samples_etvae}
\end{figure}

\begin{figure*}[!htbp]
    \centering
    \includegraphics[width=0.98\textwidth]{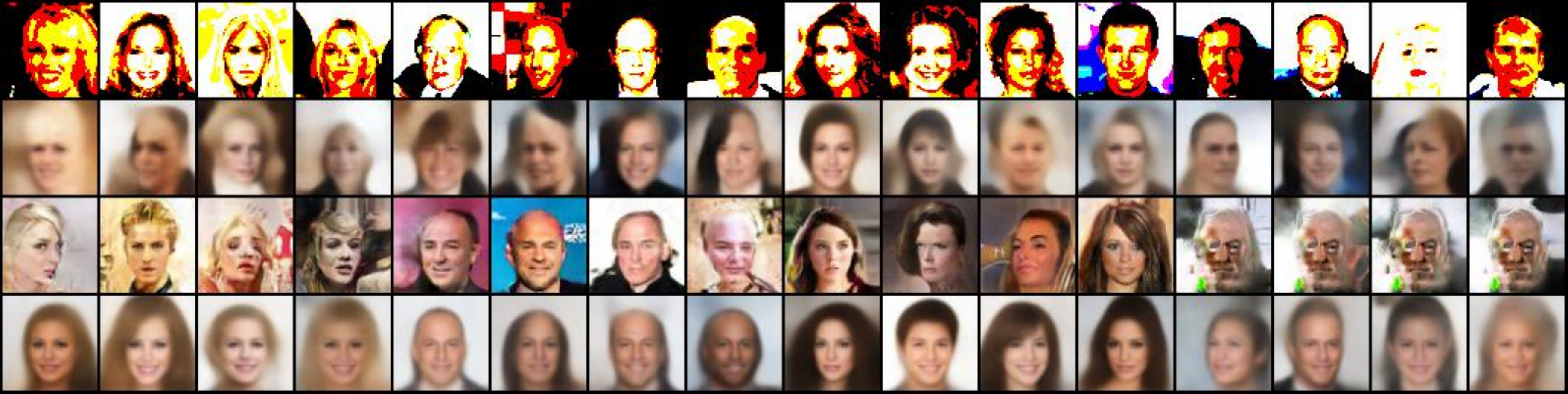}
    \caption{Comparison of the generation of different methods. Each row corresponds to one method, from top to bottom: SMOTE, OCVAE, OCDCGAN, and MGVAE. Each group of the four columns corresponds to one minority class, from left to right: blonde hair, bald hair, brown hair, and gray hair.}
    \label{fig:comparison_celeba}
\end{figure*}

\paragraph{Tabular Data.}
Finally, we test our model on several real-world tabular datasets, namely Musk, Water Quality, and Isolet. All these datasets are naturally imbalanced, as detailed in Table~\ref{tab:data_info}. For compatibility with the generative model, we preprocess all the tabular data to the scale of [-1,1] or [0,1] by dimension according to the sign of the original data. We train a 2-layer fully-connected neural network for 100 epochs with mini-batch size 100, and the learning rate is 1$e$-3. The EWC regularization weight of MGVAE is 500. 

The results of the three datasets are summarized in Tables~\ref{tab:musk}, \ref{tab:water_quality}, and \ref{tab:isolet}, respectively. Our method outperforms the other methods in all metrics, proving the effectiveness of our model on real-world data. 

\subsubsection{Qualitative Results}
To further demonstrate the effectiveness of our proposed model, we also visualize the generated minority samples that are desirably supposed to achieve the diversity and richness of the majority ones while maintaining self-semantic information.


Firstly, we visualize the generated minority samples of our model in different datasets according to the generation process of MGVAE~(see Algorithm~\ref{MGVAE_G}). The results are shown in Figure~\ref{fig:exemplar_samples_etvae}. In overall, MGVAE can generate meaningful and distinguishable samples for different datasets and imbalance ratios, i.e., no model collapse. Since the generation process of MGVAE is in a one-to-one mapping style from majority to minority, we show both the reference majority sample and the corresponding minority one. In each plate of MNIST and FashionMNIST in Figure~\ref{fig:exemplar_samples_etvae}, the left column is the reference majority sample, and the right columns are the generated minority ones correspondingly. For CelebA, the first column is the reference majority sample from the black hair. The other columns are the corresponding generated samples of the minority classes, i.e., from left to right, blond hair, bald, brown hair, and gray hair. 


Form Figure~\ref{fig:exemplar_samples_etvae}, we notice that the generated minority samples have great stylistic similarity with their reference majority samples, while the semantic information is different. For example, the number writing style in MNIST, such as tilt angle and thickness, remains the same. Some features other than hair color in CelebA, such as the face orientation and mouth shape, are consistent. In other words, the minority augmentation samples inherit the diversity of the majority ones in the process of generation. Moreover, we can see that the quality of the generated images improves as the scale of the minority increases, i.e., the imbalance ratio decreases.

Secondly, We compare the sampling quality of all over-sampling-based methods, and the results on CelebA are shown in Figure~\ref{fig:comparison_celeba}. More results on MNIST and FashionMNIST are presented in the supplementary materials~\ref{appsec:visual} because of the space constraints. As shown in Figure~\ref{fig:comparison_celeba}, we find that the samples generated by SMOTE differ greatly from the original dataset, and the sampled images of OCVAE are very blurry and have no obvious semantic features. OCDCGAN could create sharper but severely distorted images, and the model collapses when the sample size is small~(See the last four columns in the OCDCGAN row). Compared with the above methods, MGVAE can maintain the semantic information of each class while generating clear images without model collapse.

\subsection{Ablation study}
To gain a deeper understanding of the role played by each part of our model, we next perform a set of ablation experiments. 
Specifically, we first replace the prior with Gaussian to verify the effectiveness of the majority-based prior of MGVAE, i.e., we test on the standard VAE with the pretrain-finetuning training process and the EWC regularization. Second, we incrementally remove the EWC regularization and the pretrain-finetuning training process in MGVAE. The corresponding results in MNIST are summarised in Table~\ref{tab:ablation}, from which we know that these two parts can significantly improve the model's effectiveness, especially the pretrain-finetuning process. The reason is that they can avoid overfitting and model collapse by better incorporating and preserving the information of the majority class. 
\begin{table}[!htbp]
\caption{Ablation study of the classification performance in MNIST.}
\label{tab:ablation}
\begin{center}
\resizebox{\linewidth}{!}{
    \begin{tabular}{c|cll|cll}
    \toprule[2pt]
    IR & \multicolumn{3}{c|}{$\rho = 100$} & \multicolumn{3}{c}{$\rho=600$} \\ \hline
    Methods         & \multicolumn{1}{c}{B-ACC} & \multicolumn{1}{c}{ASCA} & \multicolumn{1}{c|}{GM} & \multicolumn{1}{c}{B-ACC} & \multicolumn{1}{c}{ASCA} & \multicolumn{1}{c}{GM} \\ \hline

    VAE w/ PT+EWC & \multicolumn{1}{c}{$79.1\pm 0.4$} & \multicolumn{1}{c}{$78.4 \pm 0.6$} & \multicolumn{1}{c|}{$75.7 \pm 0.4$} & \multicolumn{1}{c}{$60.3 \pm 0.8$} & \multicolumn{1}{c}{$59.3 \pm 0.7$} & \multicolumn{1}{c}{$42.7 \pm 0.3$}\\

    MGVAE w/o PT & \multicolumn{1}{c}{$81.1\pm 0.6$} & \multicolumn{1}{c}{$80.6 \pm 0.5$} & \multicolumn{1}{c|}{$78.1 \pm 0.7$} & \multicolumn{1}{c}{$59.8 \pm 0.8$} & \multicolumn{1}{c}{$58.8 \pm 0.5$} & \multicolumn{1}{c}{$41.1 \pm 1.0$}\\
    
    MGVAE w/o EWC & \multicolumn{1}{c}{$84.0\pm 0.7$} & \multicolumn{1}{c}{$83.5 \pm 0.5$} & \multicolumn{1}{c|}{$81.9 \pm 0.9$} & \multicolumn{1}{c}{$62.5 \pm 0.7$} & \multicolumn{1}{c}{$61.5 \pm 0.4$} & \multicolumn{1}{c}{$47.7 \pm 1.2$}\\
    
    \textbf{MGVAE}  & \multicolumn{1}{c}{$\textbf{85.0} \pm 0.2$} & \multicolumn{1}{c}{$\textbf{84.6} \pm 0.2$} & \multicolumn{1}{c|}{$\textbf{83.2} \pm 0.2$} & \multicolumn{1}{c}{$\textbf{65.4} \pm 1.0$} & \multicolumn{1}{c}{$\textbf{64.4} \pm 1.1$} & \multicolumn{1}{c}{$\textbf{53.4} \pm 1.1$}\\
    \bottomrule
    \end{tabular}
}
\end{center}

\end{table}

\subsection{Sensitivity Analysis on \texorpdfstring{$\lambda$}{lambda}}
In our model, we employ Elastic Weight Consolidation (EWC) regularization during the fine-tuning process to prevent catastrophic forgetting of the majority and overfitting to the minority. The corresponding hyper-parameter is $\lambda$. To analyze the impact of $\lambda$, we conduct experiments on FashionMNIST with two different imbalanced ratios. The range of values of $\lambda$ is $\{ 5e2,5e4,5e6,5e8 \}$. And the results are summarized in Table~\ref{tab:lambda}, and each averaged from three random trials.

\begin{table}[!htbp]
\caption{Comparison of classification performance of MGVAE with different $\lambda$ on FashionMNIST.}
\label{tab:lambda}
\begin{center}
\resizebox{\linewidth}{!}{
    \begin{tabular}{c|cll|cll}
    \toprule[2pt]
    IR & \multicolumn{3}{c|}{$\rho = 100$} & \multicolumn{3}{c}{$\rho=600$} \\ \hline
    $\lambda$         & \multicolumn{1}{c}{B-ACC} & \multicolumn{1}{c}{ASCA} & \multicolumn{1}{c|}{GM} & \multicolumn{1}{c}{B-ACC} & \multicolumn{1}{c}{ASCA} & \multicolumn{1}{c}{GM} \\ \hline
    
    5e2 & \multicolumn{1}{c}{$88.3 \pm 0.1$} & \multicolumn{1}{c}{$88.3 \pm 0.1$} & \multicolumn{1}{c|}{$87.5 \pm 0.1$} & \multicolumn{1}{c}{$84.1 \pm 0.6$} & \multicolumn{1}{c}{$84.1 \pm 0.5$} & \multicolumn{1}{c}{$82.8 \pm 0.4$}\\
    
    5e4  & \multicolumn{1}{c}{$88.3 \pm 0.1$} & \multicolumn{1}{c}{$88.4 \pm 0.1$} & \multicolumn{1}{c|}{$87.6 \pm 0.1$} & \multicolumn{1}{c}{$84.8 \pm 0.4$} & \multicolumn{1}{c}{$84.8 \pm 0.4$} & \multicolumn{1}{c}{$83.6 \pm 0.4$}\\
    
    5e6  & \multicolumn{1}{c}{$88.0 \pm 0.3$} & \multicolumn{1}{c}{$88.1 \pm 0.2$} & \multicolumn{1}{c|}{$87.2 \pm 0.3$} & \multicolumn{1}{c}{$84.0 \pm 0.5$} & \multicolumn{1}{c}{$84.0 \pm 0.5$} & \multicolumn{1}{c}{$82.4 \pm 0.7$}\\
    
    5e8  & \multicolumn{1}{c}{$87.7 \pm 0.3$} & \multicolumn{1}{c}{$87.7 \pm 0.3$} & \multicolumn{1}{c|}{$87.0 \pm 0.5$} & \multicolumn{1}{c}{$83.8\pm 0.6$} & \multicolumn{1}{c}{$83.8\pm 0.6$} & \multicolumn{1}{c}{$82.0 \pm 0.9$} \\
    \bottomrule
    \end{tabular}
}
\end{center}
\end{table}

\section{Related Work}
Over-sampling has been widely used to solve class-imbalanced issues with the emergence of many advanced methods in recent decades. Generally, over-sampling methods provide a way to augment the minority class information, resulting in a class-balanced dataset for the downstream tasks. Random over-sampling~(ROS)~\citep{japkowicz2000class} is one of the straightforward ways to re-balance class by repeatedly sampling the minority samples. However, ROS does not inherently have any information augmentation and is prone to overfitting in the minority classes. SMOTE~\citep{chawla2002smote} augments the minority class with new instances generated by interpolating neighboring minority class instances to address this issue. After that, some following work~\citep{han2005borderline, he2008adasyn, mullick2019generative} based on SMOTE was proposed to improve the performance. However, the synthesized samples are usually noisy due to the boundary samples, especially for image data. Besides, the performance drops drastically under the extreme imbalance ratio. Regardless of ROS or SMOTE-based methods, only instances of the minority class are used in data augmentation, which is small-scale and easy to overfit. Unlike these, major-to-minor translation~(M2m)~\citep{kim2020m2m} augments minority classes by translating the majority ones according to the trained classifier under imbalanced data, which is essentially the generation process of adversarial examples. Therefore, the augmented minority sample generated by M2m is a slight perturbation of the majority one, which is visually contrary to human cognition. More recently, CMO~\citep{park2022majority} augments the minority by leveraging the rich context of the majority classes as background images, achieved by CutMix, while the generated minority sample is semantically meaningless. Another research line of over-sampling is related to deep generative models~(DGMs), such as Conditional VAE~\citep{fajardo2021oversampling}, Contrastive VAE~\citep{dai2019generative}, GAN~\citep{pourreza2021g2d, mullick2019generative}, etc. These models can generate corresponding minority samples based on the class label. But, the model will collapse on minority class when the sample scale tends to a few. Our proposed model, MGVAE, is also a generative model, inheriting the architecture of VAE. Specifically, MGVAE generates the minority sample according to a majority-based prior, resulting in one-to-one sample mapping. The generated samples are meaningful, and class information is consistent with human cognition.

\section{Conclusion}
In the paper, we introduce Majority-Guided VAE~(MGVAE), a novel over-sampling model to re-balance datasets by generating the new minority samples based on the majority prior. By utilizing the information of the majority class, MGVAE can generate the minority samples with better diversity and richness. As a result, overfitting can be relatively avoided in downstream tasks based on the augmented data. Additionally, to better control the training process of the model, we use pretrain-finetune two-stage training and Elastic Weight Consolidation~(EWC) regularization. Finally,  we demonstrate the promising performance of the MGVAE in the downstream classification task on both benchmark image datasets and real-world tabular data.

\section*{Limitation and Societal impact}
At present, MGVAE still has some limitations. First, MGVAE is essentially a variant of VAE. Although the generation quality is much improved compared to the standard VAE, the generated images are still blurred compared to other advanced generation models such as Normalizing Flows, Diffusion models, etc. Second, in the current form of MGVAE, only the majority data is used during the pre-train process, resulting in a majority-based prior for our minority generative. We would like to test the different settings for pre-train data selection in our future work. For potential societal impact, MGVAE may be used to generate fake pictures that are not present in reality, such as human faces.

\subsubsection*{Acknowledgements}
This work was partially supported by the National Key Research and Development Program of China~(No. 2018AAA0100204), a key program of fundamental research from Shenzhen Science and Technology Innovation Commission~(No. JCYJ20200109113403826), the Major Key Project of PCL~(No. PCL2021A06), an Open Research Project of Zhejiang Lab~(NO.2022RC0AB04), and Guangdong Provincial Key Laboratory of Novel Security Intelligence Technologies~(No. 2022B1212010005).

\bibliography{ref}

\appendix
\onecolumn

\section{Derivation of Eq.(\ref{objective})}
\label{der_obj}
Following the definition in Section Methods~\ref{sec:methods}, we have
\begin{align}
    \notag
    \log&\ p(\x^- \mid \mathcal{X}^+,\Psi) \\
    \notag
    &= \log \sum_{n=1}^{N^+} \frac{1}{N^+} T_{\Psi}(\x^- \mid \x^+_n) \\
    \notag
    &= \log \sum_{n=1}^{N^+} \frac{1}{N^+} \int_{z} r_{\phi}(\z \mid \x_n^+)p_{\theta}(\x^- \mid \z) \mathbf{d}\z \\
    \notag
    &= \log \int_{z} p_{\theta}(\x^- \mid \z) \sum_{n=1}^{N^+} \frac{1}{N^+} r_{\phi}(\z \mid \x_n^+) \mathbf{d}\z \\
    \notag
    &= \log \int_{z} \frac{q_{\phi}(\z \mid \x^-)p_{\theta}(\x^- \mid \z) \sum_{n=1}^{N^+} r_{\phi}(\z \mid \x_n^+)/{N^+}}{q_{\phi}(\z \mid \x^-)} \mathbf{d}\z \\
    \notag
    &= \log \mathbb{E}_{q_{\phi}(\z \mid \x^-)} \left[p_{\theta}(\x^- \mid \z) \frac{\sum_{n=1}^{N^+} r_{\phi}(\z \mid \x_n^+)/{N^+}}{q_{\phi}(\z \mid \x^-)}\right] \\
    \notag
    &\ge \mathbb{E}_{q_{\phi}(\z \mid \x^-)} \log \left[p_{\theta}(\x^- \mid \z) \frac{\sum_{n=1}^{N^+} r_{\phi}(\z \mid \x_n^+)/{N^+}}{q_{\phi}(\z \mid \x^-)}\right] \\
    \notag
    &= \mathbb{E}_{q_{\phi}(\z \mid \x^-)} \log p_{\theta}(\x^- \mid \z) -  \log \frac{q_{\phi}(\z \mid \x^-)}{\sum_{n=1}^{N^+} r_{\phi}(\z \mid \x_n^+)/{N^+}} \\
    \notag
    &= \mathbb{E}_{q_{\phi}(\z \mid \x^-)} \log p_{\theta}(\x^- \mid \z) -\mathbb{E}_{q_{\phi}(\z \mid \x^-)} \log \frac{q_{\phi}(\z \mid \x^-)}{\sum_{n=1}^{N^+} r_{\phi}(\z \mid \x_n^+) / {N^+}} \\
    \notag
    &\equiv O(\Psi, \mathcal{X}^{+}; \x^-),
\end{align}
where the inequality relation is derived from the Jensen’s inequality.

\section{Loss re-weighting strategies}
We compare MGVAE against other different loss re-weighting strategies in the experimental section, namely FOCAL and LDAM loss. For the sake of completeness of the content, we briefly introduce these loss re-weighting strategies here.

\subsection{FOCAL Loss}
To address the imbalance problem in object detection, the Focal loss is introduced to balance the sample-wise classification loss for model training by down-weighing easy samples. In detail, the Focal loss adds a re-weighting factor $(1-h_i)^{\gamma}$ with $\gamma > 0$ to the standard cross-entropy loss $\mathcal{L}_{CE}$, where $h_i$ is a probability prediction for the sample $x_i$ over its true category $y_i$. The resultant loss takes the form as follows:
\begin{align}
    \mathcal{L}_{Focal} = (1-h_i)^{\gamma} \mathcal{L}_{CE}=-(1-h_i)^{\gamma}\log (h_i).
\end{align}

As a result, the cross-entropy loss of the easy samples, which may dominate the training by the large predicted probability $h_i$ for their true categories, will be down-weighted. Note that the $\gamma$ is set to 1.0 in all our experiments.

\subsection{LDAM Loss}
The label-distribution-aware margin~(LDAM) loss expands the decision boundaries of few-shot classes, resulting in larger margins between those classes. The final loss is formulated as a cross-entropy loss with enforced margins:
\begin{align}
    \mathcal{L}_{\text {LDAM }}:=-\log \frac{e^{\hat{y}_j-\Delta_j}}{e^{\hat{y}_j-\Delta_j}+\sum_c \neq j e^{\hat{y}_c-\Delta_c}},
\end{align}
where $\hat{y}$ are the logits and $\Delta_j$ is a class-aware margin, inversely proportional to $n_j^{1/4}$, and $n$ is the number of class samples.

\section{Detailed Training Process}
\label{appsec:training process}
To better understand our method, we summarize the detailed training process in Algorithm~\ref{alg:training_process}.
\begin{algorithm}[htbp]
    \caption{The Training Process of MGVAE.}
    \label{alg:training_process}
    \SetKwInOut{Input}{Input}
    \SetKwInOut{Output}{Output}
    \Input{Training class-imbalanced data $\mathcal{D}_{imb}=\{\mathcal{X}^+ \equiv \{\x_n^+\}_{n=1}^{N^+}, \mathcal{X}^- \equiv \{\x_n^-\}_{n=1}^{N^-}$, where $N^+ \gg N^-$. \\ 
            Batch size $B$. \\
            Pre-train steps $T_1$ and fine-tune steps $T_2$. \\
            Majority down-sample size $S$. \\
            Decoder $p_{\theta}$, encoder $q_{\phi}$, and the prior $r_{\phi}$.}
    \Output{Parameters $\theta$ and $\phi$ of generative model $M$ for minority.}
    
    \tcc{\textbf{Pre-training}}
        \For{$t=1,\cdots,T_1$}{
            \text{Obtain a mini-batch of $B$ majority datapoints} \\
            \text{Randomly down-sample $S$ majority samples $\{\x_n\}_{n=1}^{S}$for computing prior $r_{\phi}$}
            $\mathcal{L_{\text{pretrain}}} = \frac{1}{B}\sum_{i=1}^{B} \mathbb{E}_{q_{\phi}(\z \mid \x_i)} \left [\log p_{\theta}(\x_i \mid \z) - \mathbb{E}_{q_{\phi}(\z \mid \x_i)} \log \frac{q_{\phi}(\z \mid \x_i)}{\sum_{n=1}^{S} r_{\phi}(\z \mid \x_n) / {S}} \right ]$ \\
            \text{Optimize $\mathcal{L_{\text{pretrain}}}$ with ADAM optimizer.}
        }
        \text{Get pre-trained model $\mathcal{M_{\text{pretrain}}}$.} \\

    \tcc{\textbf{Fine-tuning}}
        \For{$t=1,\cdots,T_2$}{
            \text{Obtain a mini-batch of $B$ minority datapoints} \\
            \text{Randomly down-sample $S$ majority samples $\{\x_n\}_{n=1}^{S}$for computing prior $r_{\phi}$}
            \text{Get Fisher Information vector $F$ of model $M_{pretrained}$ by Eq.~\ref{eq:fisher}}
            \text{Optimize optimization objective $O_{EWC}$ in Eq.~\ref{eq:final} with ADAM optimizer.}
        }
        \text{Get trained model $\mathcal{M}$.}
\end{algorithm}

\section{More experimental results}
\subsection{More visual results}
\label{appsec:visual}
\begin{figure*}[!htbp]
    \centering
    \subfigure[MNIST~(left: $\rho=600$; right: $\rho=100$)]{
        \begin{minipage}{0.174\linewidth}
            \centering
                \includegraphics[width=0.48\linewidth]{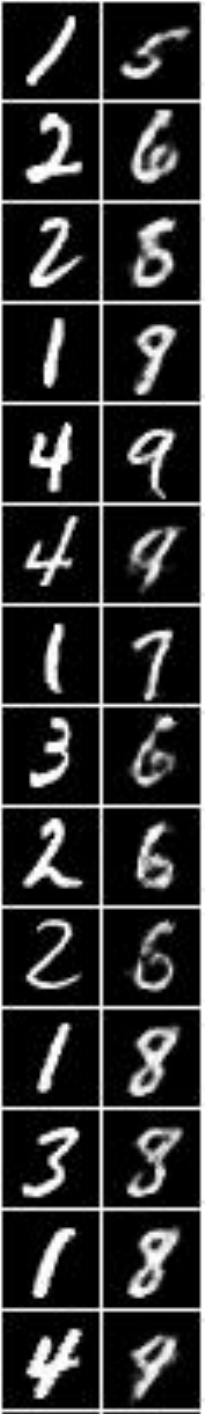}
                \includegraphics[width=0.48\linewidth]{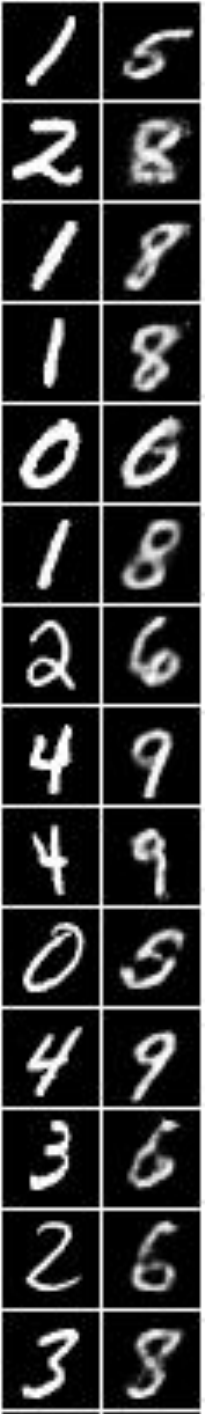}
        \end{minipage}
    }
    \subfigure[FashionMNIST~(left: $\rho=600$; right: $\rho=100$)]{
        \begin{minipage}{0.174\linewidth}
            \centering
                \includegraphics[width=0.48\linewidth]{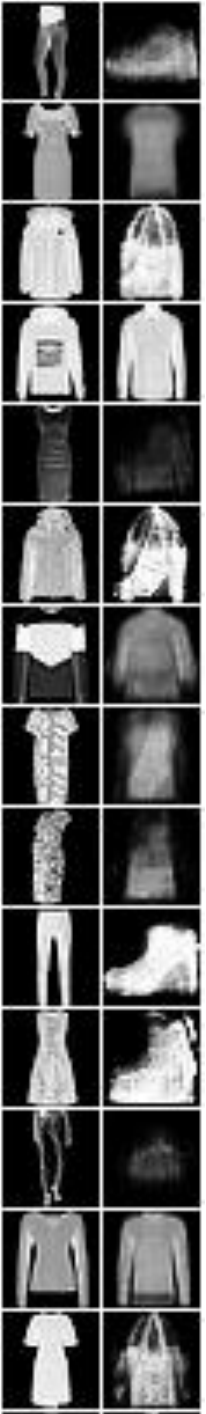}
                \includegraphics[width=0.48\linewidth]{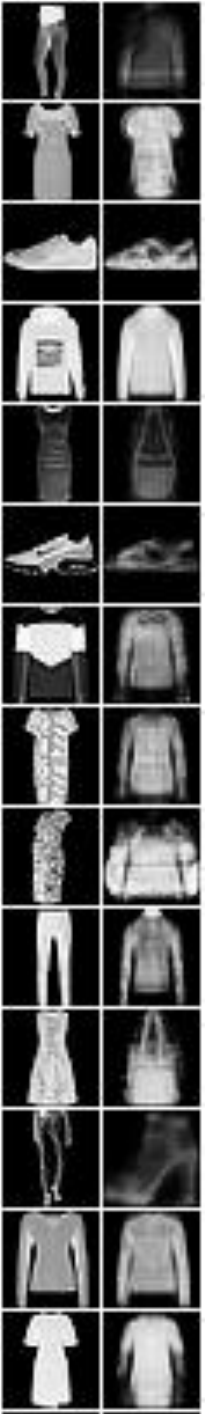}
        \end{minipage}
    }
    \subfigure[CelebA~(from left to right: black,blonde,bald,brown,and gray)]{
        \begin{minipage}{0.6\linewidth}
            \centering
                \includegraphics[width=0.48\linewidth]{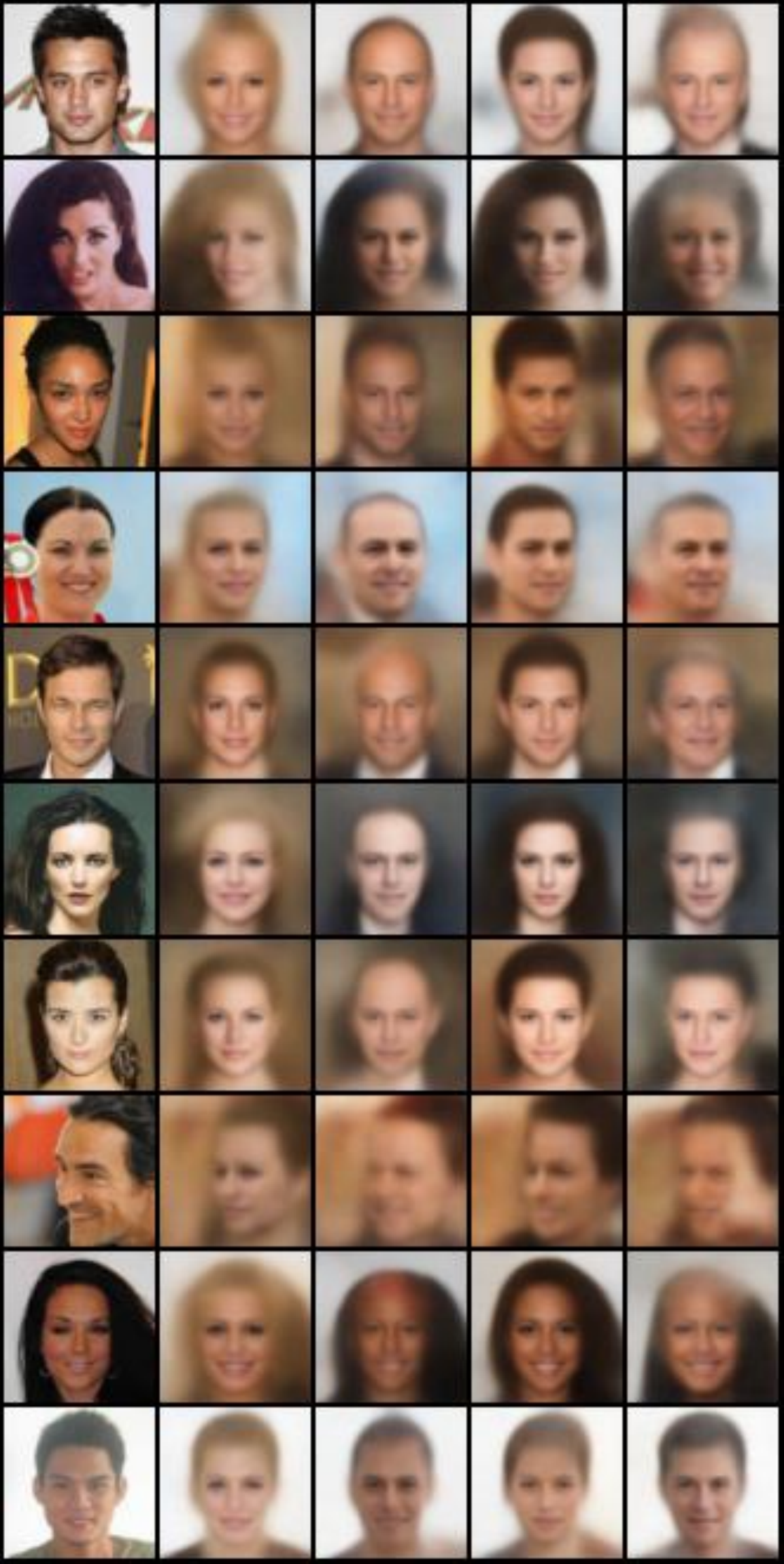}
                \includegraphics[width=0.48\linewidth]{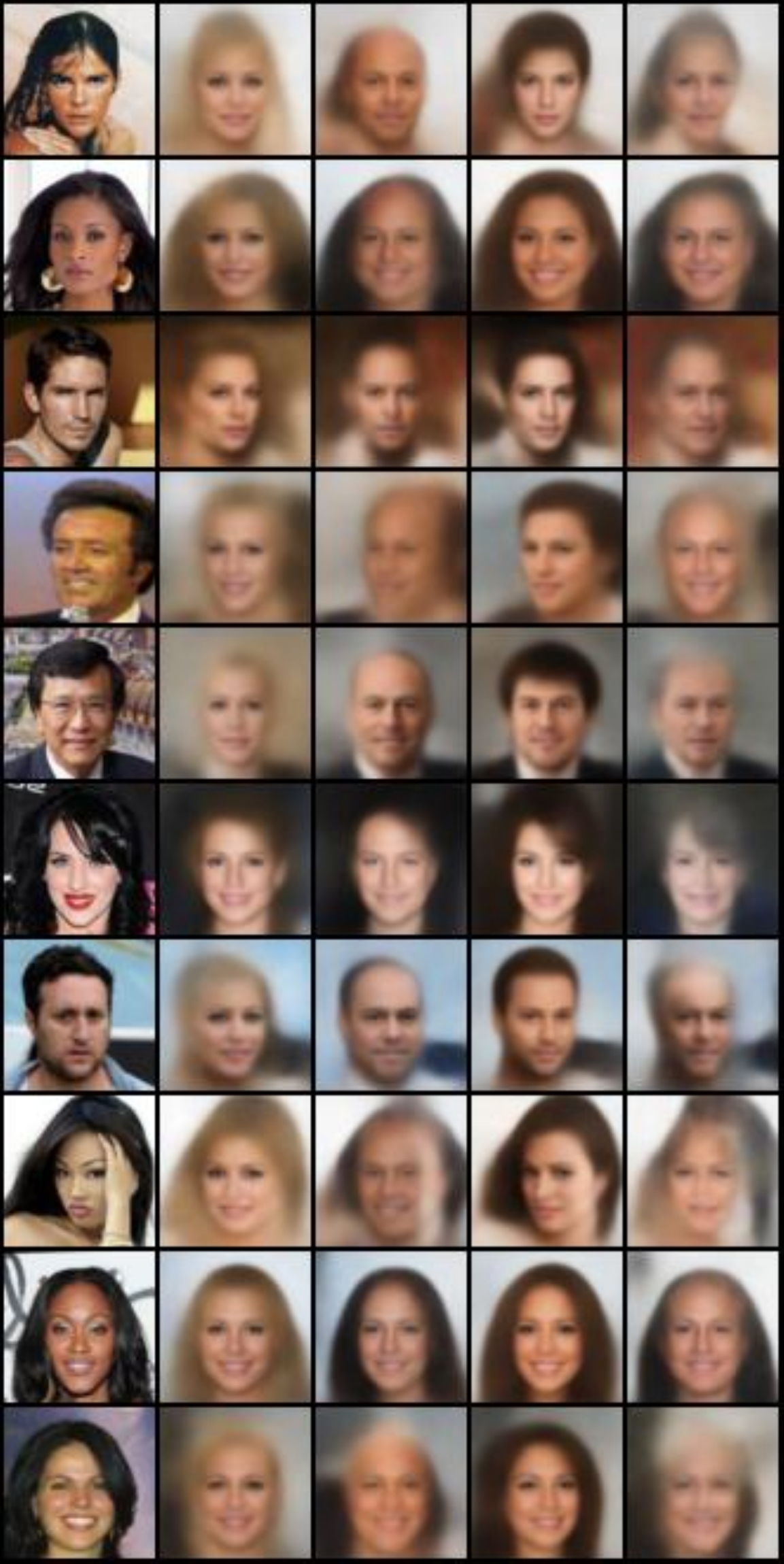}
        \end{minipage}
    }
    \caption{Samples from MGVAE.}
    \label{fig:appendix_mgvae}
\end{figure*}

Due to space constraints, we give part of the generative samples in the Qualitative Results of the experimental section. To demonstrate the effectiveness of MGVAE, we provide more visual results. 
\paragraph{Samples from MGVAEs} First, more sampling results on different models and datasets are presented in Figure~\ref{fig:appendix_mgvae}. Consistent with the main body, the left column of each plate of MNIST and FashionMNIST is the exemplar majority sample, and the right columns are the generated minority ones correspondingly. For Celeba, the first column is the exemplar majority sample from the black hair. The other columns are the corresponding generated samples of the minority classes, from left to right, blond hair, bald, brown hair, and gray hair.

\begin{figure*}[!htbp]
    \centering
    \subfigure[MNIST~(top: $\rho=600$; bottom: $\rho=100$)]{
        \begin{minipage}{0.48\linewidth}
            \centering
                \includegraphics[width=\linewidth]{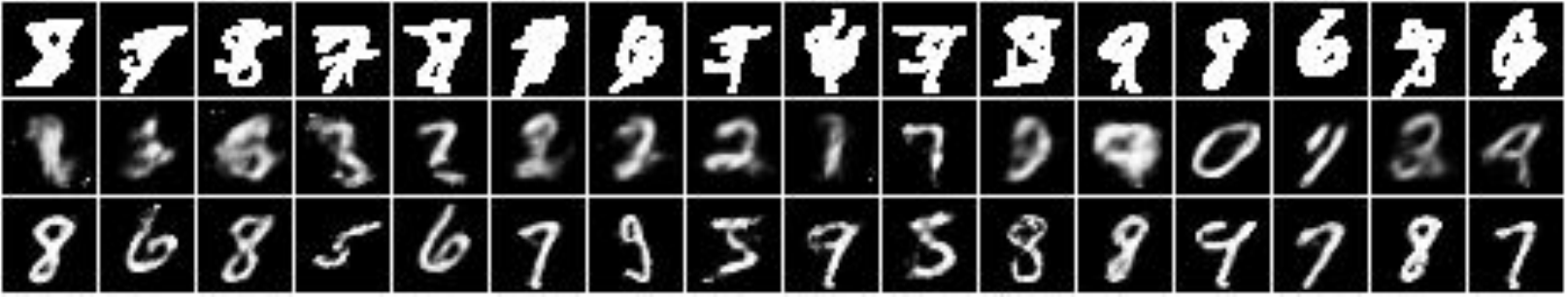}
                \includegraphics[width=\linewidth]{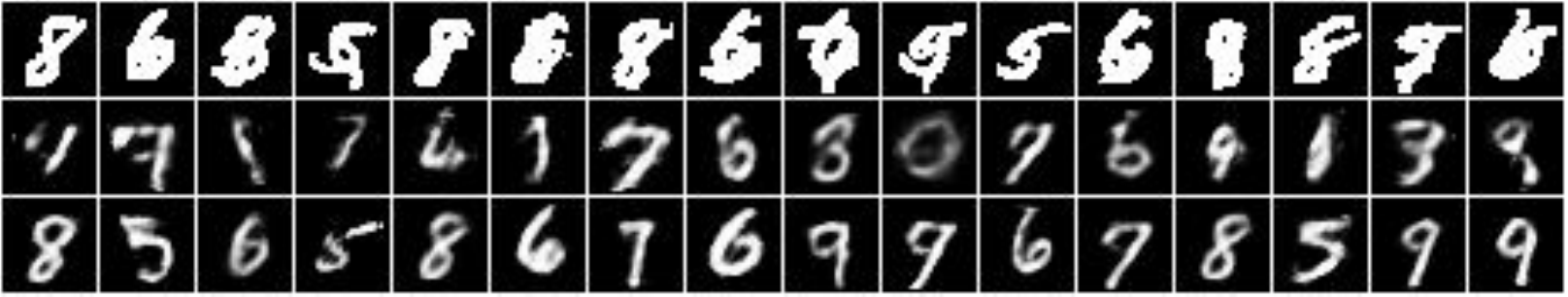}
        \end{minipage}
    }
    \subfigure[FashionMNIST~(top: $\rho=600$; bottom: $\rho=100$)]{
        \begin{minipage}{0.48\linewidth}
            \centering
                \includegraphics[width=\linewidth]{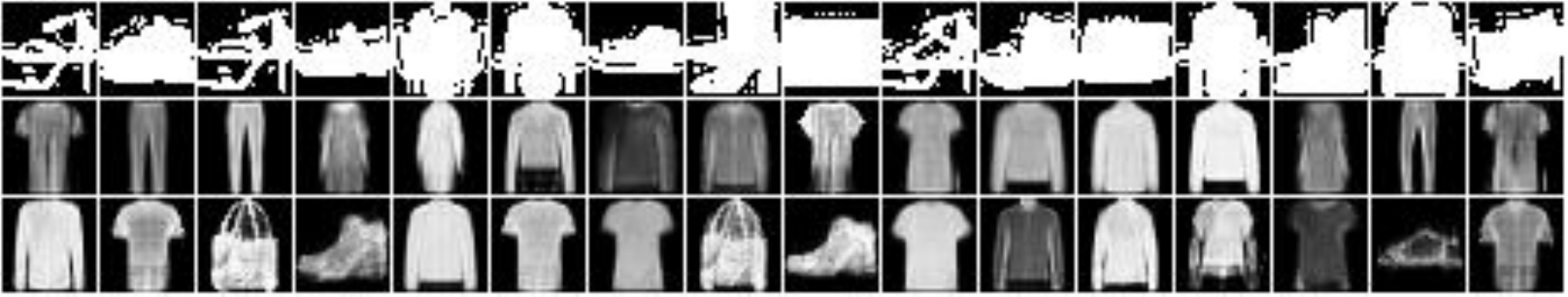}
                \includegraphics[width=\linewidth]{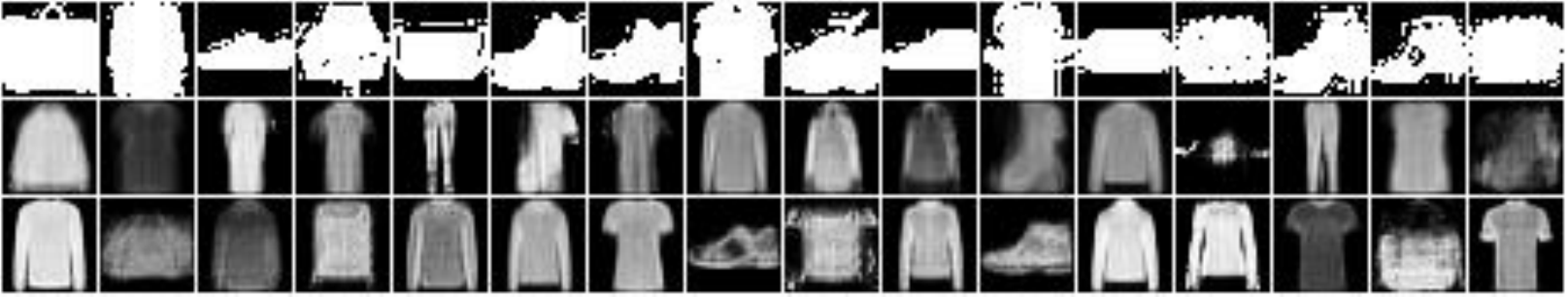}
        \end{minipage}
    }
    
    \subfigure[CelebA~(from left to right: black,blonde,bald,brown,and gray)]{
        \includegraphics[width=0.98\linewidth]{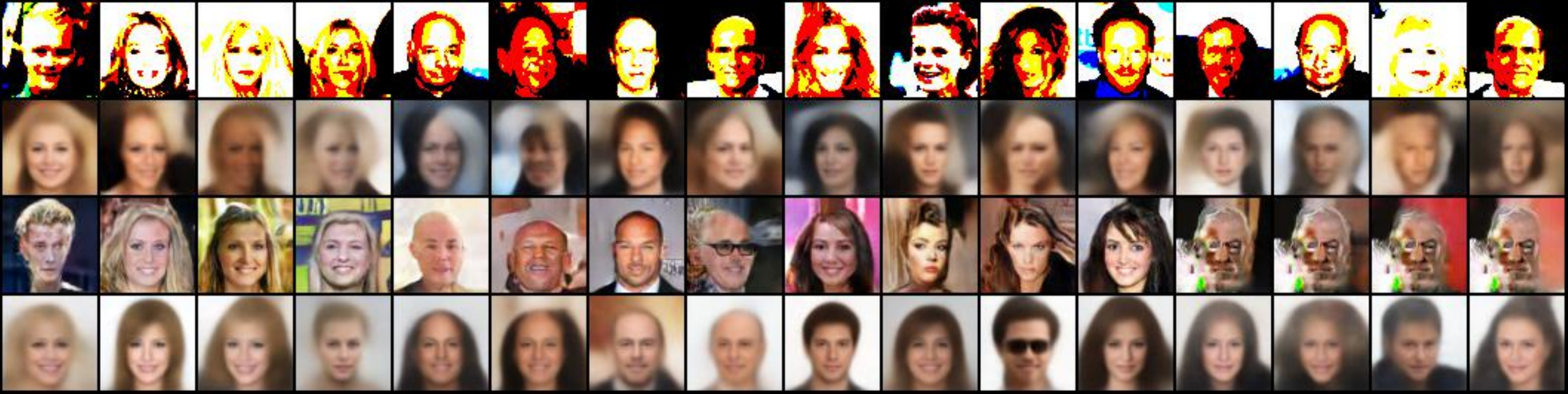}
    }
    \caption{Comparison of the generation of different methods. Each row corresponds to one method. For MNIST and FashionMNIST, from top to bottom: SMOTE, OCVAE, and MGVAE. For Celeba, from top to bottom: SMOTE, OCVAE, OCDCGAN, and MGVAE. Each group of the four columns corresponds to one minority class, from left to right: blonde hair, bald hair, brown hair, and gray hair.}
    \label{fig:appendix_com}
\end{figure*}

\paragraph{Comparison results} In the main body of the paper, we only give the comparison sampling results of different methods on Celeba due to space constraints. More results are shown in Figure~\ref{fig:appendix_com}. Compared with the other methods, MGVAE can maintain the semantic information of each class while generating clear images without model collapse.

\newpage
\subsection{Upper bounds}
For experimental completeness, we provide the results of the used classifiers on the fully-balanced MNIST and FashionMNIST, which can be considered the classification upper bounds. The results are shown in Table~\ref{tab:upper_bound}.
\begin{table}[htbp]
\caption{Classification results on fully-balanced MNIST and FashionMNIST.}
\label{tab:upper_bound}
\centering
\begin{tabular}{cll|cll}
\toprule[2pt]
 \multicolumn{3}{c|}{MNIST} & \multicolumn{3}{c}{FashionMNIST} \\ \hline
 \multicolumn{1}{c}{B-ACC} & \multicolumn{1}{c}{ASCA} & \multicolumn{1}{c|}{GM} & \multicolumn{1}{c}{B-ACC} & \multicolumn{1}{c}{ASCA} & \multicolumn{1}{c}{GM} \\ \hline
 \multicolumn{1}{c}{$98.5\pm 0.0$} & \multicolumn{1}{c}{$98.5\pm 0.0$} & \multicolumn{1}{c|}{$98.5\pm 0.0$} & \multicolumn{1}{c}{$94.3 \pm 0.1$} & \multicolumn{1}{c}{$94.3 \pm 0.1$} & \multicolumn{1}{c}{$94.2 \pm 0.1$}\\
\bottomrule
\end{tabular}
\end{table}

\section{Architectures}
\subsection{Architectures of the DGMs}
For a fair comparison, the structure of all generated models is kept consistent for each particular dataset. Specifically, the neural network uses fully-connected layers~(denoted MLP) for the relatively simple datasets MNIST, FashionMNIST, and Tabular data. For Celeba, a convolutional layer~(denoted CNN) is used. Note that the architecture of the convolutional layer is based on this code repository\footnote{\url{https://github.com/sajadn/Exemplar-VAE/blob/master/models/fully_conv.py}}. We use curly brackets to denote concatenation; the number in a bracket means the layer size.
\paragraph{VAE-based architecture with MLP:} including MGVAE and OCVAE for MNIST, FashionMNIST, and Tabular data.
\begin{align}
\textbf{Encoder:} \notag\\
    \text{E1} &= \text{MLP}\ [\text{Input dim}-\text{Hidden dim 1}] \notag\\
    \text{E2} &= \text{MLP}\ [\text{Hidden dim 1}-\text{Hidden dim 2}] \notag \\
    \log \sigma^2 &= \text{MLP}\ [\text{Hidden dim 2}-\text{Latent dim}] \notag \\
    \mu_z &= \text{MLP}\ [\text{Hidden dim 2}-\text{Latent dim}] \notag \\
\textbf{Decoder:} \notag \\
    \text{D1} &= \text{MLP}\ [\text{Latent dim}-\text{Hidden dim 2}] \notag \\
    \text{D2} &= \text{MLP}\ [\text{Hidden dim 2}-\text{Hidden dim 1}] \notag \\
    \mu_x &= \text{MLP}\ [\text{Hidden dim 2}-\text{Output dim}] \notag
\end{align}
Specifically, for MNIST and FashionMNIST, Input dim=Output dim=784, Hidden dim 1=Hidden dim 2=300, and Latent dim=40. For Tabular data, Input dim=Output dim=Data dim, Hidden dim 1=Hidden dim 2=300, and Latent dim=10.

\paragraph{VAE-based architecture with CNN:} including MGVAE and OCVAE for Celeba.
\begin{align}
\textbf{Encoder:} \notag\\
    \text{E1} &= \text{CNN}\ [\text{64 × 64 × 3}-\text{32 × 32 × 64}] \notag\\
    \text{E2} &= \text{CNN}\ [\text{32 × 32 × 64}-\text{16 × 16 × 128}] \notag \\
    \text{E3} &= \text{CNN}\ [\text{16 × 16 × 128}-\text{8 × 8 × 256}] \notag \\
    \text{E4} &= \text{CNN}\ [\text{8 × 8 × 256}-\text{4 × 4 × 512}] \notag \\
    \log \sigma^2 &= \text{MLP}\ [\text{4 × 4 × 512}-\text{Latent dim}] \notag \\
    \mu_z &= \text{MLP}\ [\text{4 × 4 × 512}-\text{Latent dim}] \notag \\
\textbf{Decoder:} \notag \\
    \text{D1} &= \text{MLP}\ [\text{Latent dim}-\text{4 × 4 × 512}] \notag \\
    &\text{Upsample(2)} \notag\\
    \text{D2} &= \text{CNN}\ [\text{8 × 8 × 512}-\text{16 × 16 × 256}] \notag \\
    \text{D3} &= \text{CNN}\ [\text{16 × 16 × 256}-\text{32 × 32 × 128}] \notag \\
    \text{D4} &= \text{CNN}\ [\text{32 × 32 × 128}-\text{64 × 64 × 64}] \notag \\
    \mu_x &= \text{CNN}\ [\text{64 × 64 × 64}-\text{64 × 64 × 3}] \notag 
\end{align}
We use Latent dim=40 for all our models in Celeba, including MGVAE and OCVAE.

\paragraph{Architecture of OCDCGAN:} including OCDCGAN for Celeba.
\begin{align}
\textbf{Generator:} \notag\\
    \text{G1} &= \text{CNN}\ [\text{Latent dim} - \text{4 × 4 × 1024}] \notag\\
    \text{G2} &= \text{CNN}\ [\text{4 × 4 × 1024}-\text{8 × 8 × 512}] \notag\\
    \text{G3} &= \text{CNN}\ [\text{8 × 8 × 512}-\text{16 × 16 × 256}] \notag \\
    \text{G4} &= \text{CNN}\ [\text{16 × 16 × 256}-\text{32 × 32 × 128}] \notag \\
    \text{G5} &= \text{CNN}\ [\text{32 × 32 × 128}-\text{64 × 64 × 3}] \notag \\
\textbf{Discriminator:} \notag \\
    \text{G1} &= \text{CNN}\ [\text{64 × 64 × 3} - \text{32 × 32 × 128}] \notag\\
    \text{G2} &= \text{CNN}\ [\text{32 × 32 × 128} - \text{16 × 16 × 256}] \notag\\
    \text{G3} &= \text{CNN}\ [\text{16 × 16 × 256} - \text{8 × 8 × 512}] \notag\\
    \text{G4} &= \text{CNN}\ [\text{8 × 8 × 512} - \text{4 × 4 × 1024}] \notag\\
    \text{G5} &= \text{CNN}\ [\text{4 × 4 × 1024} - \text{1 × 1 × 1}] \notag\\
    &\text{Sigmoid()} \notag
\end{align}
The Latent dim is 100 in our setting.
\subsection{Architectures of the Classifiers}
Similarly, for the classifier, we also use two network structures: fully-connected layers for MNIST, FashionMNIST, Tabular data, and ResNet-20 for Celeba. The architecture of the fully-connected network is as follows. See our code implementation for details.
\paragraph{MLP classifier.}
\begin{align}
    \text{C1} &= \text{MLP}\ [\text{Input dim}-\text{256}] \notag\\
    \text{C2} &= \text{MLP}\ [\text{256}-\text{128}] \notag \\
    \text{C3} &= \text{MLP}\ [\text{128}-\text{Class num}] \notag 
\end{align}
The architecture of the ResNet-20 network is based on this implementation\footnote{\url{https://github.com/akamaster/pytorch_resnet_cifar10/blob/master/resnet.py}}.

\vfill

\end{document}